\pdfoutput=1
\documentclass[11pt]{article}

\usepackage[preprint]{acl}

\usepackage{times}
\usepackage{latexsym}

\usepackage[T1]{fontenc}
\usepackage[utf8]{inputenc}
\usepackage{makecell}

\usepackage{microtype}

\usepackage{inconsolata}

\usepackage{graphicx}

\usepackage{times}
\usepackage{latexsym}
\usepackage{times}
\usepackage{latexsym}
\usepackage{booktabs}
\usepackage{boxedminipage}
\usepackage{amssymb}
\usepackage{amsmath}
\usepackage{todonotes} 
\usepackage{graphicx}

\usepackage{hyperref}
\usepackage{inconsolata}
\usepackage{pifont}% http://ctan.org/pkg/pifont

\usepackage{tabu}
\usepackage{fontawesome}
\usepackage{amsmath}
\usepackage{amsfonts}
\usepackage{amssymb}
\usepackage{enumitem}
\usepackage{listings}
\usepackage{xstring}
\usepackage{graphicx}
\usepackage{pbox}
\usepackage{subcaption}
\usepackage{epstopdf}
\usepackage{xstring}
\usepackage{multirow}

\usepackage{soul}
\usepackage{color}
\usepackage{graphicx}
\usepackage{multirow}
\usepackage{comment}
\usepackage{listings} % Add this for using the listings package
\usepackage{algorithm}
\usepackage{algpseudocode}

\usepackage{graphicx}
\usepackage{subcaption}
\usepackage{verbatim}

\usepackage{amssymb}
\usepackage{xcolor} % For coloring text
\definecolor{lightblue}{rgb}{.50,.90,0.51}
\definecolor{tri}{rgb}{.25,.88,.82}
\definecolor{lilac}{rgb}{0.85,0.64,0.85}
\definecolor{atomictangerine}{rgb}{1.0, 0.6, 0.4}
\usepackage[table]{xcolor}

\newcommand{\dsname}{\textsc{ProBel}}
\newcommand{\ourbest}{\textsc{Mt-Sft}}

\usepackage{tabularx}

\title{\textsf{\textbf{ProBel}}: Propaganda Detection with Techniques, Spans, and Explanations}

\author{Mohamed Bayan Kmainasi$^1$, Ali Ezzat Shahroor$^1$, Elisa Sartori$^2$, \\
{\bf  Giovanni Da San Martino$^2$, Firoj Alam$^1$}\\
  $^1$Qatar Computing Research Institute, Qatar, 
  $^2$University of Padova, Italy \\
  {\tt \{mkmainasi, alsh34060,fialam\}@hbku.edu.qa},\\ 
  {\tt elisa.sartori.7@studenti.unipd.it, giovanni.dasanmartino@unipd.it}\\
  {\small \tt \href{https://huggingface.co/collections/QCRI/media-integrity-intelligence}{https://huggingface.co/collections/QCRI/media-integrity-intelligence}}
\\}

\begin{document}
\maketitle
\begin{abstract}
Propaganda detection includes several related prediction levels, ranging from sentence-level decisions to technique classification and span identification. However, it remains unclear how supervision at these levels interacts when learned jointly across Arabic and English. We present \textit{ProBel} an Arabic and English resource that aligns binary labels, multi-label annotations over 23 propaganda techniques grouped into six coarse categories, technique-labeled spans, and reference explanations for the same news sentences. It includes a substantially larger English collection and supports matched binary, coarse-grained, multi-label, and span-level tasks in both languages. We evaluate zero-shot prompting, task-specific fine-tuning, and joint training under a shared setup. A single \textit{bilingual multi-task model achieves the best overall performance} and remains competitive across tasks and languages. Cross-task analysis shows that transfer depends on the supervision level. Joint classification training preserves binary performance, whereas span-only training can weaken sentence-level prediction. Joint bilingual training yields the most stable results, while monolingual fine-tuning can reduce transfer to the other language. We will release the data, code, and evaluation scripts.\footnote{
Resources:
\href{https://github.com/MohamedBayan/ProBel}{Code}
|
\href{https://huggingface.co/datasets/QCRI/ProBel}{Dataset}
|
\href{https://huggingface.co/QCRI/ProBel-MTL}{Model}
}

\end{abstract}

\section{Introduction}
\label{sec:intro}

Protecting public information has become an urgent global concern. The World Economic Forum ranks misinformation and disinformation as the second most severe global risk over the next two years \cite{wef_grr2026}. Propaganda contributes to this challenge by shaping public opinion through selective framing, emotional appeals, and other manipulative strategies \cite{barron2019proppy,DaSanMartino2019emnlp}. Timely detection can help readers, journalists, fact-checkers, and moderators examine such content before it spreads widely.

Propaganda detection has been studied through sentence-level decisions, fine-grained technique classification, and span identification \cite{DaSanMartino2019emnlp,DaSanMartino2020semeval,piskorski2023semeval,hasanain-etal-2024-gpt}. These tasks provide complementary views of the same content. However, they are often developed and evaluated separately. It remains unclear how supervision transfers across these tasks and whether the effects are consistent across Arabic and English.

\begin{figure*}[t!]
\centering
\includegraphics[width=0.9\textwidth]{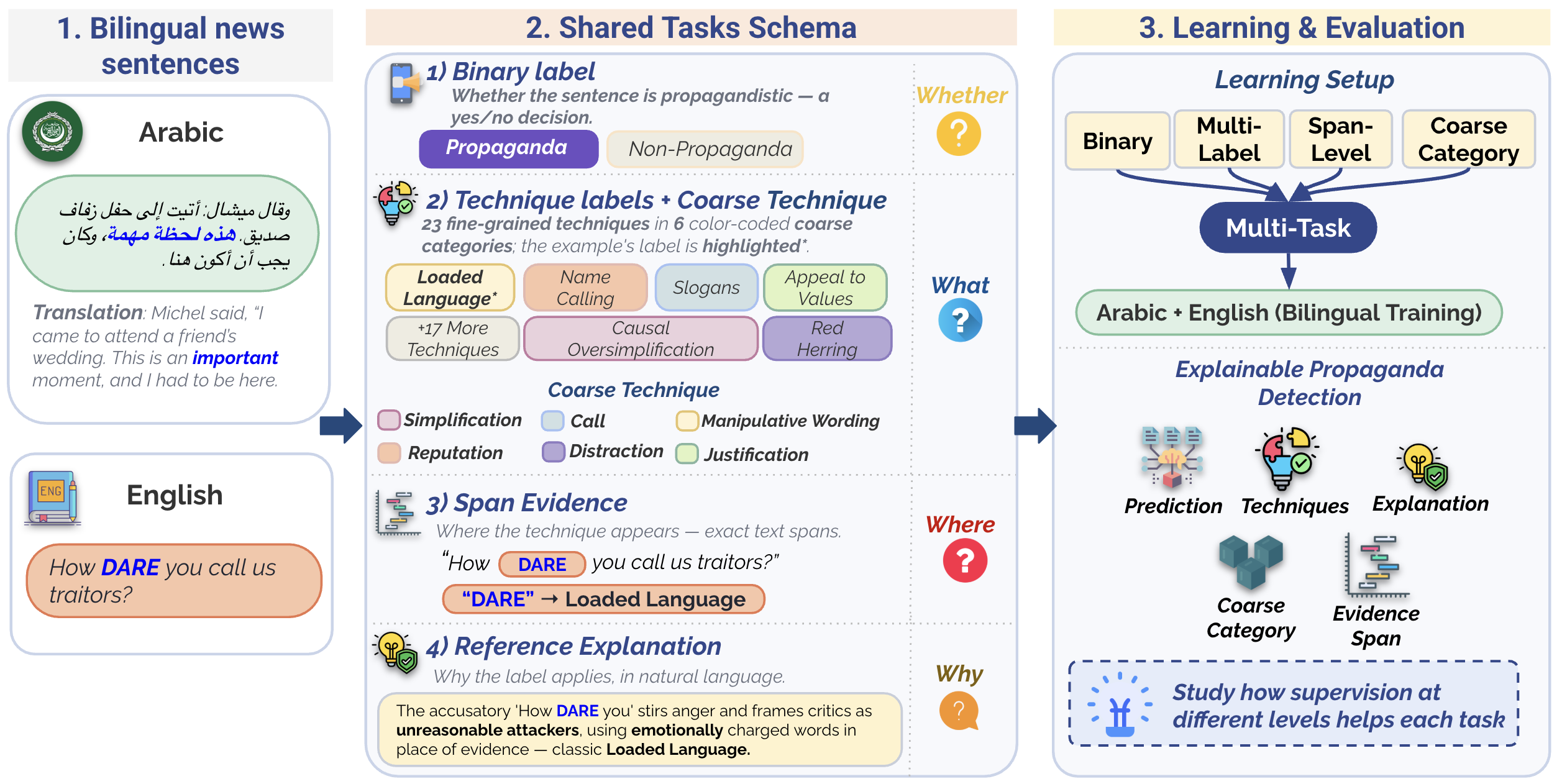}
\vspace{-0.2cm}
\caption{
Overview of the aligned tasks and learning setup. Each Arabic and English sentence includes a binary label, fine-grained techniques grouped into six categories, labeled evidence spans, and a reference explanation, supporting task-specific and joint learning.
% Overview of the aligned annotation and learning setup. Each Arabic and English sentence is annotated with a binary label, fine-grained technique labels grouped into six coarse categories, technique-labeled evidence spans, and a reference explanation, enabling task-specific and joint learning across tasks.
}
\label{fig:overview}
\vspace{-0.3cm}
\end{figure*}

As shown in Figure~\ref{fig:overview}, propaganda analysis can address four interrelated questions. \textit{Is} propaganda present, \textit{which} technique is used, \textit{where} does it appear, and \textit{why} does the identified text support the decision? Binary labels, multi-label technique annotations, technique-labeled spans, and natural language explanations address these questions, respectively. These tasks are related because a positive decision should be supported by an identified technique and corresponding textual evidence.

Existing resources cover parts of this structure. SemEval-2020 studies propaganda spans and techniques in English news \cite{DaSanMartino2020semeval}, SemEval-2023 studies persuasion techniques in multilingual news \cite{piskorski2023semeval}, and ArPro provides fine-grained Arabic span annotations \cite{hasanain-etal-2024-gpt}. PropXplain pairs Arabic and English binary labels with reference explanations \cite{hasanain2025propxplain}. These resources separate annotation tasks across datasets, languages, or evaluation settings, limiting controlled analysis of cross-task supervision.

We present \dsname, an Arabic–English resource aligning binary labels, multi-label annotations for 23 propaganda techniques, technique-labeled spans, and reference explanations for news sentences and social media posts. It substantially expands the English collection while supporting the same tasks and taxonomy in both languages. This design enables controlled comparisons across task-specific, multi-task, monolingual, and bilingual settings, and allows us to assess whether tasks reinforce one another or cause negative transfer during joint training.

We organize the study around three research questions.

\begin{description}[noitemsep,topsep=0pt,labelsep=.5em]
    \item[RQ1] How do zero-shot prompting, in-context learning, task-specific fine-tuning, and multi-task learning perform across binary detection, technique classification, and span identification?
    
    \item[RQ2] How does supervision transfer across binary, technique-level, and span-level tasks during joint training?
    
    \item[RQ3] How do the expanded English data and joint Arabic and English training affect performance across languages?

    \item[RQ4] Can distillation, from a stronger teacher or from the model's own privileged predictions, replace direct fine-tuning?

\end{description}

Our results show that a bilingual multi-task model performs best overall and remains competitive with specialized models across tasks and languages. Transfer varies by supervision level: joint classification preserves binary performance, whereas span-only training can weaken sentence-level prediction. Joint Arabic–English training yields the most stable results and avoids the degradation observed with monolingual fine-tuning. Our contributions are as follows.

% Our results show that a single bilingual multi-task model achieves the strongest overall performance and remains competitive with specialized models across tasks and languages. Transfer depends on the supervision level. Joint classification training preserves binary performance, whereas span-only training can weaken sentence-level prediction. Joint Arabic and English training provides the most stable results and avoids the cross-lingual degradation observed after monolingual fine-tuning. Our contributions includes.

\begin{itemize}[noitemsep,topsep=0pt,labelsep=.5em]
    \item \textbf{A unified bilingual resource.} \dsname aligns binary labels, 23-technique multi-label annotations, technique-labeled spans, and reference explanations for Arabic and English content.
    
    \item \textbf{An expanded English collection.} The English data are substantially enlarged and support the same annotation levels as the Arabic data, while the Arabic splits remain comparable with earlier results.
    
    \item \textbf{A controlled cross-task study.} Task-specific and multi-task models are evaluated under a shared setup to measure transfer across annotation levels.
    
    \item \textbf{A bilingual transfer analysis.} Monolingual and joint Arabic and English training are compared across all core tasks, together with analyses of cross-language transfer and long-tailed techniques.

    \item \textbf{A distillation analysis.} We show that neither teacher traces nor privileged self-distillation matches direct fine-tuning on \dsname, diagnose why the published on-policy recipe stalls on our task, and repair it with a saturating clip.    
    \end{itemize}

\section{Related Work}
\label{sec:related_work}

Research on propaganda detection spans sentence-level classification, technique identification, and span detection \cite{DaSanMartino2019emnlp,DaSanMartino2020semeval}. These tasks have also been extended to multilingual and multimodal settings \cite{SemEval2021-6-Dimitrov,piskorski2023semeval,dimitrov2024semeval}. Recent work also investigates extractive rationales and natural language explanations to make model decisions easier to inspect \cite{camburu2018esnli,mathew2021hatexplain,wiegreffe2021teach,hase2022when}.

\subsection{Propaganda detection across tasks}

Early work examined article-level propaganda identification and organization \cite{barron2019proppy}. Da San Martino et al.\ introduced 18 propaganda techniques for sentence-level binary detection and fragment-level technique identification \cite{DaSanMartino2019emnlp}. SemEval-2020 later evaluated span identification and technique classification in English news using 14 consolidated techniques \cite{DaSanMartino2020semeval}.

Subsequent shared tasks expanded language, domain, and modality coverage. SemEval-2021 studied propaganda techniques in English memes \cite{SemEval2021-6-Dimitrov}, SemEval-2023 evaluated 23 persuasion techniques in multilingual news \cite{piskorski2023semeval}, and SemEval-2024 addressed multilingual memes using a hierarchy of 22 techniques \cite{dimitrov2024semeval}. For Arabic, ArPro provides news paragraphs annotated with 23 techniques at paragraph and span levels \cite{hasanain-etal-2024-gpt}.

These resources offer complementary binary, technique, and span annotations. However, differences in task coverage, language, annotation design, and evaluation settings hinder controlled analysis of supervision transfer across tasks and between Arabic and English.

\subsection{Span-Level Rationales} 
Span-level rationales highlight input segments that support a model prediction and improve interpretability in tasks such as fact-checking, hate speech, and deception detection \cite{mathew2021hatexplain,yu2021interpretable,russo2023benchmarking}. In propaganda detection, annotated spans identify text expressing a technique, while natural language rationales justify the assigned label. Recent work has explored generating such explanations \cite{10.1145/3613904.3642805,atanasova2024generating,hasanain2025propxplain}. However, few studies jointly examine propaganda labels, technique spans, and natural language explanations in a unified setting.

\subsection{Comparison with Prior Resources}

\textit{PropXplain}~\cite{hasanain2025propxplain} is the closest resource to \dsname, providing Arabic and English propaganda instances with binary labels and reference explanations. However, \dsname has broader task coverage, aligning binary labels, 23 fine-grained techniques, technique-labeled spans, and natural language rationales. It also expands the English collection and supports controlled analysis of cross-task supervision, multilingual transfer, and joint Arabic-English training.
In Table~\ref{tab:resource-comparison}, we compare the task coverage of related resources. Earlier English datasets introduced binary, technique, and span annotations, while later shared tasks expanded their linguistic and multimodal scope. ArPro provides binary, technique, and span annotations for Arabic, whereas \textit{PropXplain} adds explanations for Arabic and English binary classification. \dsname uniquely combines bilingual data with all four annotation levels under a shared 23-technique taxonomy.

% \begin{table}[t]
% \centering
% \scriptsize
% \setlength{\tabcolsep}{2.0pt}
% % \renewcommand{\arraystretch}{1.08}
% \scalebox{0.75}{
% \begin{tabular}{@{}p{5.0cm}ccccc@{}}
% \toprule
% Resource & Lang. & Bin. & Tech. & Span & NLE \\
% \midrule
% Da San Martino et al. \cite{DaSanMartino2019emnlp} & EN & \checkmark & 18 & \checkmark & -- \\
% SemEval-2020 \cite{DaSanMartino2020semeval} & EN & -- & 14 & \checkmark & -- \\
% SemEval-2023 \cite{piskorski2023semeval} & 9L & -- & 23 & -- & -- \\
% ArPro \cite{hasanain2024arpro} & AR & \checkmark & 23 & \checkmark & -- \\
% SemEval-2024 \cite{dimitrov2024semeval} & 4L & \checkmark & 22 & -- & -- \\
% PropXplain \cite{hasanain2025propxplain} & AR/EN & \checkmark & -- & -- & \checkmark \\
% \midrule
% \textbf{\dsname} & AR/EN & \checkmark & 23 & \checkmark & \checkmark \\
% \bottomrule
% \end{tabular}
% }
% \vspace{-0.2cm}
% \caption{Task coverage in prior propaganda resources. \textbf{Bin.:} binary sentence- or paragraph-level classification; \textbf{Tech.}, the number of fine-grained techniques; \textbf{Span:}, technique-annotated text spans; and NLE, natural language explanations. 9L: nine languages.}
% \label{tab:resource-comparison}
% \vspace{-0.2cm}
% \end{table}

\section{\dsname{} Resource}
\label{sec:dataset}

\subsection{Overview}

\begin{figure}[!tbh]
\centering
\includegraphics[width=0.96\columnwidth]{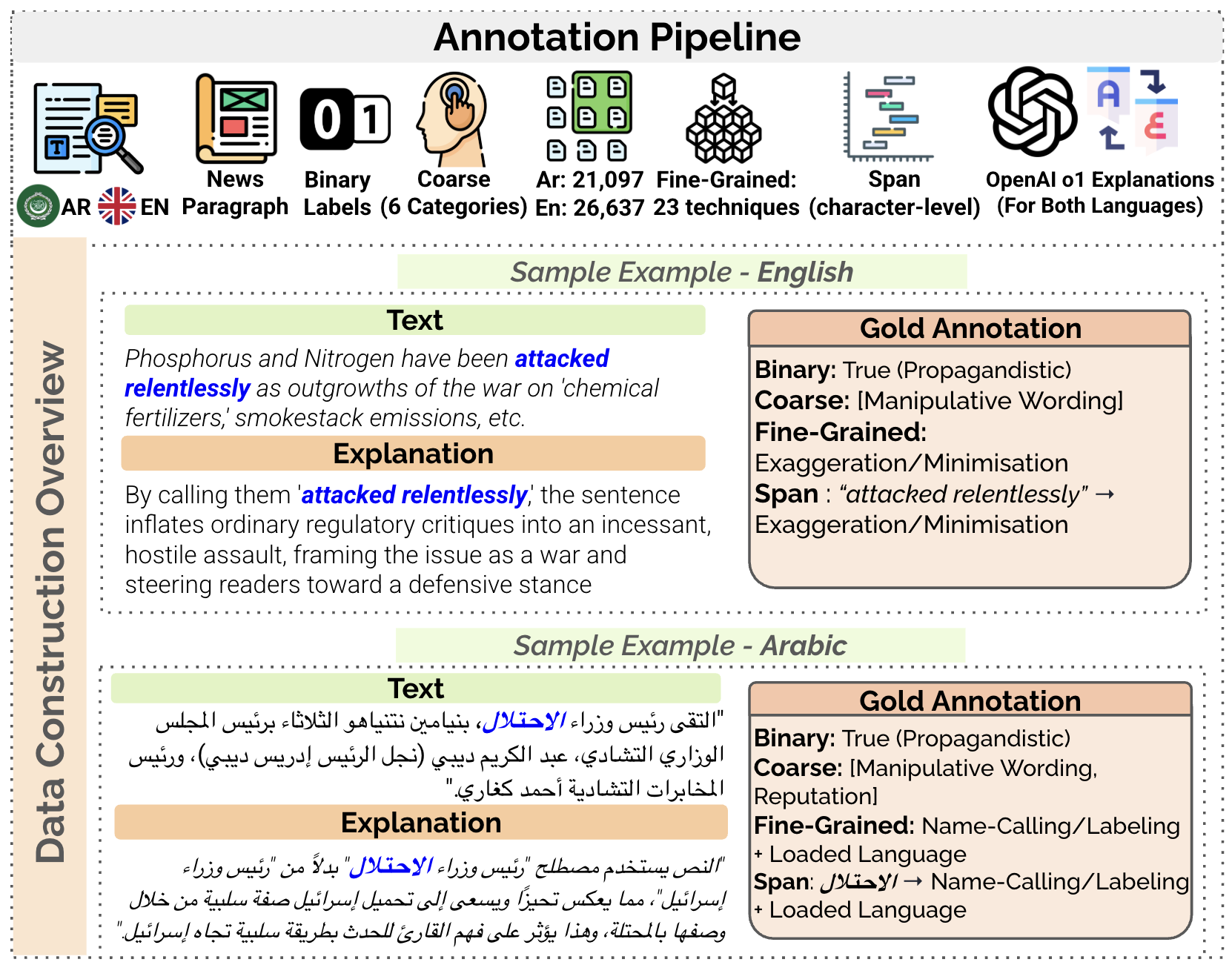}
\vspace{-0.2cm}
\caption{Overview of the construction of \dsname{}. Arabic and English content is represented with a binary label, six coarse categories, 23 fine-grained techniques, annotated text spans, and a reference explanation. 
% The examples illustrate the common annotation structure.
}
\label{fig:pipeline}
\vspace{-0.3cm}
\end{figure}

\dsname{} is an Arabic and English resource for explainable propaganda detection. Each instance includes four complementary forms of supervision. These consist of a binary propaganda label, one or more labels from a shared inventory of 23 techniques, text spans annotated with their corresponding techniques, and a reference natural language explanation. The resource contains news sentences and social media posts and supports binary classification, fine-grained multi-label classification, and span identification under a common annotation structure. Figure~\ref{fig:pipeline} summarizes the construction process.

\subsection{Source collections and splits}
\label{sec:source_data}

\noindent\textbf{Arabic data.}
The Arabic collection is collected from \textit{PropXplain} consisting of news and social media content \cite{hasanain2025propxplain}. The news portion contains paragraphs from articles published by 300 news agencies and covers 14 topics, including politics, human rights, and science and technology. The social media portion focuses on the Israeli--Palestinian war. It was collected using 14 manually selected keywords and phrases that reflected topics discussed during October and early November 2023. The Twitter search API was used to retrieve posts published during the second week of November 2023, resulting in approximately 5.7K posts. 
% We retain the existing Arabic train, development, and test splits to support comparison with prior work.

\noindent\textbf{English data.}
The English collection extends previously released data \cite{hasanain2025propxplain} with 97 additional articles, resulting in 347 articles from 42 news sources. The articles collection covers topics discussed during late 2023 and early 2024, with substantial coverage of politics and the Israeli--Palestinian war. 
% We manually removed collection artifacts, including isolated links and incorrectly retained content. The articles were then segmented into sentences.

% The data were divided into training, development, and test sets at the \textcolor{red}{[sentence/article]} level. 
% \textcolor{red}{[If the split is sentence-level, report the number or percentage of articles shared across splits here or in the limitations section.]}

\begin{table}[t]
\centering
\small
\setlength{\tabcolsep}{2.2pt}
\scalebox{0.8}{
\begin{tabular}{@{}lccp{2.35cm}@{}}
\toprule
\textbf{Lang.} &
\begin{tabular}[c]{@{}c@{}}\textbf{PropXplain}\\Train/Dev/Test\end{tabular} &
\begin{tabular}[c]{@{}c@{}}\textbf{\dsname}\\Train/Dev/Test\end{tabular} &
\textbf{Added in this work} \\
\midrule
AR &
\makecell{18,453/1,318/\\1,326} &
\makecell{18,453/1,318/\\1,326} &
Technique labels and annotated spans \\

EN &
\makecell{4,472/621/\\922} &
\makecell{18,775/2,567/\\3,993} &
Additional data, technique labels, annotated spans, and explanations \\
\bottomrule
\end{tabular}
}
\vspace{-0.2cm}
\caption{Comparison \dsname{} with PropXplain. 
% \cite{hasanain2025propxplain}. 
% PropXplain provides binary labels and reference explanations in both languages. \dsname{} retains the Arabic splits, expands the English collection, and supports fine-grained technique and span annotations in both languages.
}
\label{tab:dataset-comparison}
\vspace{-0.3cm}
\end{table}

In Table~\ref{tab:dataset-comparison}, we summarize the relation between \dsname{} and PropXplain. The Arabic splits are retained unchanged, while the English collection is substantially expanded. The present resource also supports fine-grained technique classification and span identification in both languages.

\subsection{Annotation structure}
\label{sec:annotation_structure}

% Each instance \(x\) is represented as
For the \dsname{}, as shown in Figure \ref{fig:pipeline}, we use a common structure. Each instance \(x\) is represented as
$
\mathcal{A}(x) = \langle b, Y, S, e \rangle,
$
where \(b\) is the binary propaganda label, \(Y\) is the set of fine-grained technique labels, \(S\) contains the annotated spans and their corresponding techniques, and \(e\) is the reference natural language explanation. We use the same representation for Arabic and English.

\noindent\textbf{Binary propaganda label.}
Each instance is labeled as either \textit{propagandistic} or \textit{non-propagandistic}. An instance is considered propagandistic when at least one propaganda technique is annotated and non-propagandistic otherwise. We use this rule to align the binary and fine-grained annotations \cite{hasanain-etal-2024-gpt,hasanain2025propxplain}.

\noindent\textbf{Fine-grained technique labels.}
We use the taxonomy adopted in prior multilingual and Arabic propaganda resources \cite{piskorski2023semeval,hasanain-etal-2024-gpt}. 
It organizes 23 fine-grained techniques. An instance may receive several labels when it contains more than one technique.

\noindent\textbf{Annotated text spans.}
For each occurrence of a technique, annotators identify the text span that expresses it and assign the corresponding technique label. An instance may contain several spans, and the same technique may appear more than once. 
% Spans may also overlap when the same wording expresses multiple techniques \cite{hasanain2024arpro}. 
We retain the character boundaries and technique label for every span.

\noindent\textbf{Reference explanations.}
Each instance includes a natural language explanation in the same language as the input. For propagandistic content, the explanation describes how the relevant wording expresses the annotated techniques. For non-propagandistic content, it explains why the text does not contain a propaganda technique. Following PropXplain, the explanations for the spans are generated using the gold labels, techniques and annotated spans. 
% We treat them as reference outputs rather than replacements for the structured annotations.

\subsection{Annotation}
\label{sec:data_annotation}

\noindent\textbf{Technique and span annotation.}
We retain the fine-grained annotations from the source collections and apply the 23 techniques to the added English data. Each article was annotated independently by at least two annotators. Annotators selected all applicable techniques and marked the spans that expressed them. An expert annotator reviewed the annotations and resolved disagreements. 
% The annotation team inspected random samples throughout the process and provided feedback when inconsistencies were found.

\noindent\textbf{Explanation generation.}
Following \citet{hasanain2025propxplain}, we use GPT-o1 to generate explanations from the gold binary label, technique labels, and annotated spans. Each explanation is produced in the input language, including for the newly added English data.

% We follow the procedure described by \citet{hasanain2025propxplain}, in which GPT o1 model is used for explanation generation. The generation prompt provides the gold binary label, technique labels, and annotated spans. The explanations are produced in the language of the input. We apply the same procedure to the newly added English data.
% For propagandistic instances, the model produces an explanation of how the selected wording supports the annotation. For non-propagandistic instances, it explains why no propaganda technique is present. 
% The explanations are produced in the language of the input. We apply the same procedure to the newly added English data.

\noindent\textbf{Annotation agreement.}
\label{sec:data_quality}
For Arabic, three annotators independently annotated each paragraph, and two expert consolidators reviewed the annotations~\cite{hasanain-etal-2024-gpt}. The agreement between the individual annotations and the consolidated labels using \(\gamma\) is 0.546 for span annotations. For English, the corresponding agreement score was 0.535.

% \paragraph{Explanation validation}
We generate explanations following the procedure of \citet{hasanain2025propxplain}. In that study, three annotators evaluated each explanation in the Arabic and English test sets for \textit{faithfulness}, \textit{clarity}, \textit{plausibility}, and \textit{informativeness} using a five-point scale. Agreement, measured with \(r^{*}_{wg(j)}\), ranged from 0.89 to 0.92 for Arabic and from 0.94 to 0.95 for English. To limit additional annotation cost, we do not repeat the human evaluation for the additional English set.

\subsection{Dataset statistics}
\label{sec:dataset_statistics}

In Table~\ref{tab:dataset-composition}, we report the split-wise data distribution. The dataset contains $\sim$48K instances, including $\sim$21 Arabic and $\sim$27K English instances. 
% We use $\sim$39K instances for training, 3,885 for development, and 5,319 for testing. Arabic contributes 15,419 news sentences and 5,678 tweets, whereas English contributes 26,637 news sentences. 
Propagandistic instances account for 63.7\% of Arabic and 28.0\% of English.

\begin{table}[!tbh]
\centering
% \small
\setlength{\tabcolsep}{3pt}
\scalebox{0.8}{
\begin{tabular}{@{}lrrrrr@{}}
\toprule
\textbf{Lang.} & \textbf{Train} & \textbf{Dev} & \textbf{Test} & \textbf{Total} & \textbf{Pos. (\%)} \\
\midrule
AR & 18,453 & 1,318 & 1,326 & 21,097 & 63.7 \\
EN & 20,077 & 2,567 & 3,993 & 26,637 & 28.0 \\ \midrule
\textbf{Total} & 38,530 & 3,885 & 5,319 & 47,734 & 43.8 \\
\bottomrule
\end{tabular}
}
\vspace{-0.2cm}
\caption{Dataset distribution by language. Pos.: the percentage of propagandistic instances.}
\label{tab:dataset-composition}
\vspace{-0.2cm}
\end{table}

In Table~\ref{tab:annotation-statistics}, we summarize text length and the number of techniques and spans per propagandistic instance. Arabic shows a more skewed distribution, with its three most frequent techniques covering 94.8\% of propagandistic instances, compared with 67.0\% in English. The full 23-technique distribution appears in the supplementary material. The mean explanation lengths for Arabic and English are 50.0 and 56.4 words, respectively.
% Explanations cover 99.94\% of Arabic instances and all English instances, with 

% \begin{table}[t]
% \centering
% % \scriptsize
% \setlength{\tabcolsep}{2.2pt}
% % \renewcommand{\arraystretch}{1.05}
% \scalebox{0.8}{
% \begin{tabular}{@{}lrrrrr@{}}
% \toprule
% \textbf{Lang.} & \textbf{Text} & \textbf{Tech.} & \textbf{Span} & \textbf{Multi-T} & \textbf{Multi-S} \\
% \midrule
% AR & 34.1 & 1.8 & 3.0 & 53.3 & 70.7 \\
% EN & 23.8 & 1.3 & 1.5 & 24.9 & 30.6 \\
% \bottomrule
% \end{tabular}
% }
% \vspace{-0.2cm}
% \caption{Annotation statistics by language. \textbf{Text:} the mean input length in words. \textbf{Tech.} and \textbf{Span} represents mean counts over propagandistic instances. \textbf{Multi-T} and \textbf{Multi-S} represents the percentages of propagandistic instances with more than one technique or span.}
% \label{tab:annotation-statistics}
% \vspace{-0.2cm}
% \end{table}

% In Figure~\ref{fig:technique-distribution}, we show the ten most frequent techniques across both languages. Loaded language ranks first in Arabic and English, followed by name calling. Arabic shows greater concentration, with its three most frequent techniques covering 94.8\% of propagandistic instances, compared with 67.0\% in English. The full 23-technique distribution appears in the supplementary material.

The technique distribution is heavily skewed: the three most frequent techniques cover 94.8\% of Arabic and 67.0\% of English propagandistic instances (Figure~\ref{fig:technique-distribution}, Appendix~\ref{app:labels}); per-technique test support is in Table~\ref{tab:pertech}.

\section{Tasks and Experimental Setup}
\label{sec:tasks}

\subsection{Task formulations}

We study four core prediction tasks as presented in Figure \ref{fig:overview}.

\noindent\textbf{Binary propaganda detection.}
Given an input text, the model predicts whether it contains at least one propaganda technique. We compute macro-F$_1$ for this task. 
% as the primary metric as it gives equal importance to both classes.
% despite differences in label prevalence across Arabic and English.

\noindent\textbf{Multilabel technique classification.}
Given an input text, the model predicts all applicable techniques from the 23-label taxonomy. We compute micro-F$_1$ to account for multiple labels. 

\noindent\textbf{Coarse-grained technique classification.}
The 23-technique taxonomy is highly imbalanced, with a few frequent techniques and limited support for many others, a common pattern in propaganda datasets \cite{hasanain-etal-2024-gpt}. To reduce label sparsity and support more reliable learning and evaluation, we map the 23 fine-grained techniques to six broader categories, as shown in Table~\ref{tab:labelmap}. 
For the multilibel classification, the task is formulated as follows. Given an input text, the model predicts all applicable coarse-grained categories. 
% Since an instance may express more than one category, we formulate the task as multi-label classification and 
We compute micro-F$_1$ for task as well.

\begin{figure}[t]
\centering
\includegraphics[width=0.85\columnwidth]{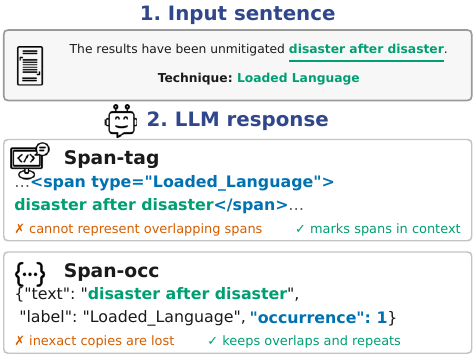}
\vspace{-0.2cm}
\caption{Two LLM span representations. \textit{Span-tag} inserts inline tags, while \textit{span-occ} outputs one object per span and supports repeated or overlapping spans.}
% \caption{The two span representations used with LLMs, on a \dsname{} sentence with one gold span. Span-tag reproduces the input with inline tags and cannot represent overlapping spans; span-occ emits one object per span, with an occurrence ordinal for repeated strings.}
\label{fig:spanformats}
\vspace{-0.4cm}
\end{figure}

% \begin{figure}[t]
% \centering
% \includegraphics[width=\columnwidth]{figures/fig_span_formats.pdf}
% \caption{The two span representations used with LLMs, on a \dsname{} sentence with one gold span. Span-tag reproduces the input with inline tags and cannot represent overlapping spans; span-occ emits one object per span, with an occurrence ordinal for repeated strings.}
% \label{fig:spanformats}
% \vspace{-0.4cm}
% \end{figure}

\noindent\textbf{Span detection with technique labels}
Given an input text, the model extracts each propagandistic span and assigns the corresponding technique. We evaluate two output representations. \textit{Span-tag} reproduces the input with inline technique tags, whereas \textit{Span-occ} returns a JSON list of \textit{(technique, span, occurrence)} triples. An example is provided in Figure \ref{fig:spanformats}. The occurrence index distinguishes repeated surface forms. We evaluate both representations using the overlap-adjusted micro-F$_1$ of \citet{DaSanMartino2020semeval}.

\noindent\textbf{Explanation generation}
For each classification formulation, generative models also produce an explanation justifying the predicted label. We compute BERTScore-F$_1$ \cite{zhang2020bertscoreevaluatingtextgeneration} to evaluate the generated explanations.
% We evaluate the generated explanations using BERTScore-F$_1$.
% and report the results separately from the core prediction tasks.

\subsection{Comparison systems}

\noindent\textbf{Baselines.}
We compute majority-label and random baselines for the classification formulations. For span detection, we use a whole-text baseline that assigns the most frequent technique to the complete input. 
% These baselines provide reference points under the imbalanced class and technique distributions.

\noindent\textbf{Prompting baselines}
We evaluate open-weight and proprietary LLMs under zero-shot prompting using only the task instructions, label definitions, and required output format. For in-context learning (ICL), we retrieve $k$ demonstrations from the training pool with BGE-M3 embeddings \cite{chen2024bgem3}; the retriever, pool, and $k$ are selected on development data (Appendix~\ref{app:icl}).

% Zero-shot prompting
% We evaluate open-weight and proprietary LLMs using only the task instructions, label definitions, and required output format. No demonstrations or task-specific parameter updates are provided.
% \noindent\textbf{In-context learning}
% We retrieve demonstrations from the corresponding training pool using BGE-M3 embeddings. The selected examples are inserted into the prompt before the test instance. We compare retrieved demonstrations with random selection and evaluate different values of \(k\).

% \noindent\textbf{Fine-tuned models}
% We compare task-specific and multi-task generative models. Task-specific fine-tuning trains a separate model for each task formulation and language setting. Multi-task fine-tuning uses one model across the all task formulations and both languages. We use Qwen2.5-7B-Instruct as the shared generative backbone and fine-tune it with LoRA \cite{hu2022lora}. We set the rank to 16, \(\alpha\) to 32, the learning rate to \(1\times10^{-5}\), the number of epochs to four, and the maximum sequence length to 4,096 tokens. We select checkpoints using development loss and keep this configuration fixed across all fine-tuning experiments.
% %
% We also evaluate discriminative baselines: sequence-classification heads on Llama-3.1-8B and Qwen2.5-7B, and fine-tuned encoders, AraBERT-v2 for Arabic and BERT-base for English (Appendix~\ref{app:baselines}).

\noindent\textbf{Fine-tuned models.}
We compare task-specific models trained separately for each task and language with a bilingual multi-task model trained across all formulations. All generative models use Qwen2.5-7B-Instruct with LoRA \cite{hu2022lora}, with checkpoints selected by development loss. We provide Hyperparameters in Appendix~\ref{app:setup}. We also evaluate discriminative baselines using classification heads on Llama-3.1-8B and Qwen2.5-7B, along with AraBERT-v2 for Arabic and BERT-base for English (Appendix~\ref{app:baselines}).

\noindent\textbf{Decoding and parsing.}
We use fixed output templates, a shared parser, and greedy decoding with zero temperature. In Appendix~\ref{app:setup}, we provide the templates, parsing details, and hyperparameters.
% \noindent\textbf{Decoding and parsing.}
% Generative systems follow a fixed output template for each formulation. A shared parser extracts labels, spans, and explanations and handles invalid or incomplete outputs consistently across systems. We use greedy decoding with temperature set to zero. Appendix~\ref{app:setup} provides the output templates and hyperparameters.

\subsection{Controlled comparisons}
\label{sec:training_regimes}

% \noindent\textbf{Cross-task transfer.}
% We compare four training setups. The first uses binary supervision only. The second uses the three classification formulations covering binary, coarse-grained, and fine-grained prediction. The third uses the two span formulations. The fourth uses the five formulations. We evaluate each setup on every formulation, including those not observed during training, to measure transfer across tasks.

% \noindent\textbf{Cross-language transfer.}
% We compare Arabic-only, English-only, and joint Arabic and English training under the same multi-task configuration. This analysis examines whether supervision in one language improves performance in the other or weakens its existing capabilities.

\noindent\textbf{Cross-task transfer.}
We compare four training setups. The first uses binary supervision only. The second uses the three classification formulations covering binary, coarse-grained, and fine-grained prediction. The third uses the two span formulations. The fourth uses the five formulations. We evaluate each setup on every formulation, including those not observed during training, to measure transfer across tasks.

\noindent\textbf{Cross-language transfer.}
We compare Arabic-only, English-only, and joint Arabic and English training under the same multi-task configuration. This analysis examines whether supervision in one language improves performance in the other or weakens its existing capabilities.

\paragraph{Distillation.}
We also test whether the aligned annotations can be replaced by cheaper supervision: off-policy distillation from GPT-5 traces \cite{hsieh2023distilling}, and on-policy self-distillation from a frozen teacher that sees the gold annotation in context \cite{zhao2026opsd}. Beyond this comparison, we diagnose why the published on-policy objective stalls and propose a saturating clip that repairs it.

\section{Experimental Results}

\subsection{Baseline Results}
\label{sec:baselines}

% We establish what binary detection looks like without task-specific training (Table~\ref{tab:zeroshot-main}). No zero-shot system reaches 0.69 macro-F$_1$ in either language. Scale and access do not decide the outcome: in each language the best open model beats GPT-5 with chain-of-thought, 0.672 against 0.646 on Arabic and 0.683 against 0.657 on English, while Gemini-3.1-Pro with chain-of-thought reaches 0.505 on Arabic, barely above the 0.499 of a fair coin. Four of the six open models fall below that chance level on Arabic, so several systems are not detecting propaganda there at all. Fine-tuning changes the picture: \ourbest{} adds 9.1 points on Arabic and 5.3 on English over the best prompted system. Exact McNemar tests on the paired predictions confirm the key comparisons of this paper at $p{<}0.001$, except GPT-5 versus Llama-3.1-8B on Arabic at $p{=}0.012$ (Appendix~\ref{app:significance}).

In Table~\ref{tab:zeroshot-main}, we show that zero-shot performance remains limited and varies substantially across models and languages. The strongest open model outperforms GPT-5 in each language, while several models approach or fall below the random baseline on Arabic. Fine-tuning yields clear gains over all prompted systems in both languages. Exact McNemar tests confirm the main improvements as statistically significant. We provide further details about the statistical significance in Appendix~\ref{app:significance}.

\begin{table}[t]
\centering
\small
\setlength{\tabcolsep}{4.5pt}
\scalebox{0.8}{
\begin{tabular}{@{}lcc@{}}
\toprule
\textbf{System} & \textbf{AR} & \textbf{EN} \\
\midrule
\multicolumn{3}{@{}l}{\emph{No learning}} \\
Majority label        & 0.380 & 0.419 \\
Random, fair coin     & 0.499 & 0.470 \\
\midrule
\multicolumn{3}{@{}l}{\emph{Zero-shot, open weights}} \\
Fanar-2-27B~\cite{fanarteam2026fanar20arabicgenerative}  & 0.409 & 0.431 \\
Llama-3.1-8B~\cite{grattafiori2024llama3herdmodels} & \underline{0.672} & 0.437 \\
Qwen2.5-7B~\cite{qwen2.5}                         & 0.439 & \underline{0.683} \\
Qwen3-VL-8B~\cite{yang2025qwen3technicalreport}           & 0.433 & 0.669 \\
Qwen3-VL-8B (think)                                       & 0.577 & 0.625 \\
Gemma-4-E4B~\cite{gemmateam2026gemma4}                    & 0.442 & 0.671 \\
\midrule
\multicolumn{3}{@{}l}{\emph{Zero-shot, proprietary}} \\
GPT-5 (CoT)~\cite{singh2026openaigpt5card}               & 0.646 & 0.657 \\
Gemini-3.1-Pro (CoT)\footnotemark                        & 0.505 & 0.673 \\
\midrule
\rowcolor{gray!15}
\textbf{\ourbest{} (fine-tuned)} & \textbf{0.763} & \textbf{0.735} \\
\bottomrule
\end{tabular}
}
\vspace{-0.2cm}
\caption{\textbf{Binary} macro-F$_1$ without task-specific training; best open model per language in bold.
% No zero-shot system, open or proprietary, reaches 0.69. The proprietary models answer one multi-task prompt covering all five tasks; direct-prompting variants, this protocol difference, and all other tasks are in Appendix~\ref{app:baselines}.
}
\label{tab:zeroshot-main}
\vspace{-0.3cm}
\end{table}
\footnotetext{\url{https://deepmind.google/models/gemini/pro/}}

\subsection{Results across Tasks \& Settings}
\label{sec:rq1}

% RQ1 covers the main task comparison.

\begin{figure}[t!]
\centering
\includegraphics[width=0.96\columnwidth]{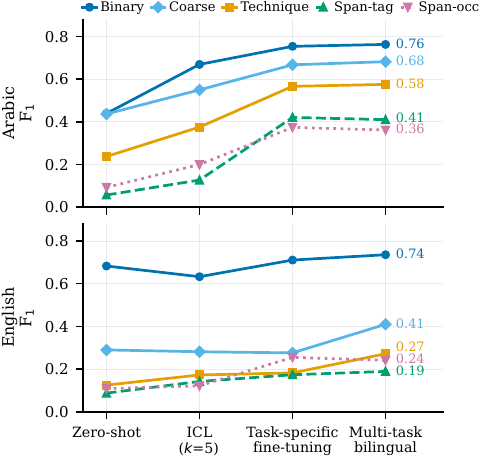}
\vspace{-0.1cm}
\caption{Performance of Qwen2.5-7B-Instruct across five tasks and four training settings. 
% \ourbest{} denotes bilingual multi-task fine-tuning. 
Full configurations and ablations appear in Appendices~\ref{app:setup} and~\ref{app:icl}.}
% \caption{Qwen2.5-7B-Instruct on all five tasks across the four training settings: zero-shot prompting, in-context learning with five BGE-M3 same-language demonstrations, task-specific fine-tuning (one model per task and language), and multi-task bilingual fine-tuning (\ourbest, one model for all five tasks in both languages). Each task uses its primary metric: binary macro-F$_1$; coarse and technique micro-F$_1$; spans overlap-adjusted micro-F$_1$. Full configurations and ablations are in Appendix~\ref{app:setup} and \ref{app:icl}.}
\label{fig:rq1}
\vspace{-0.3cm}
\end{figure}

% \noindent\textbf{Difficulty increases with output granularity.} As shown in Figure~\ref{fig:rq1}, every setting scores highest on binary classification, lower on coarse and lower again on technique classification, with the span tasks lowest overall; the one exception is span-occ under English task-specific fine-tuning (0.255), which overtakes its technique score.

\noindent\textbf{Difficulty increases with output granularity.} In Figure~\ref{fig:rq1}, we show that performance declines as outputs become more fine-grained. Models perform best on binary classification, followed by coarse and technique classification, and generally perform worst on span prediction. English task-specific \textit{span-occ} provides the only exception.

% \todo{add details in appendix later, model, few-shot, strategies, etc. and add the figure}

% \noindent\textbf{In-context learning.} Retrieved demonstrations improved performance across all Arabic tasks, e.g., binary detection increased from 0.439 to 0.669. In English, however, performance remained largely unchanged, except for binary classification, where demonstrations reduced performance because the zero-shot setting was already strong. Nevertheless, demonstrations never closed the gap on span detection, where the best score remained at or below 0.20 in both languages.

\noindent\textbf{In-context learning.} Few-shot ICL improves performance across all Arabic tasks, including binary detection. In English, it provides limited gains and reduces binary performance, where zero-shot prompting is already strong. It also fails to close the gap on span detection, which remains the most challenging task in both languages.

% \noindent\textbf{A single bilingual multi-task model performs best overall.} \ourbest{} matches or outperforms the task-specific models on every classification task and concedes at most 1.4 points on the span tasks (e.g., 0.411 vs.\ 0.421 on Arabic span tagging). This demonstrates that a single model can maintain or improve performance while jointly handling Arabic and English, predicting binary labels, techniques, spans, and explanations from a single checkpoint, thereby simplifying deployment.
% %
% English is consistently harder than Arabic on the structured tasks. Its test set contains only 27.8\% propagandistic sentences, compared with 61.3\% in Arabic, providing fewer positive examples per technique and reducing fine-grained performance by roughly half.

\noindent\textbf{A single bilingual multi-task model performs best overall.} \ourbest{} matches or outperforms task-specific models on all classification tasks and remains within 1.4 points on span prediction (See Fig \ref{fig:rq1}). It handles both languages and all tasks. English remains more challenging as its test set contains fewer propagandistic instances, which limits positive examples for fine-grained techniques.

% \noindent\textbf{The two span representations disagree across languages.} In Figure~\ref{fig:rq1}, span-tag is the stronger format on Arabic (0.421 vs.\ 0.374 for the task-specific models) and span-occ on English (0.255 vs.\ 0.173), so we report both. Table~\ref{tab:spanres} in Appendix~\ref{app:spanformats} compares all systems on the span task, where the BIO encoders surpass every generative system on English except the span-occ fine-tuned models.

\noindent\textbf{Span representations vary by language.} 
In Figure~\ref{fig:rq1}, we find that \textit{span-tag} performs better in Arabic, while \textit{span-occ} performs better in English. We therefore report both formats. In Appendix~\ref{app:spanformats}, we provide detail results and comparisons.

% \subsection{Transfer across Annotation Levels}
\subsection{Transfer across Tasks}
\label{sec:rq2}
% RQ2 covers cross-task transfer.

\begin{figure}[!tbh]
\centering
\includegraphics[width=0.96\columnwidth]{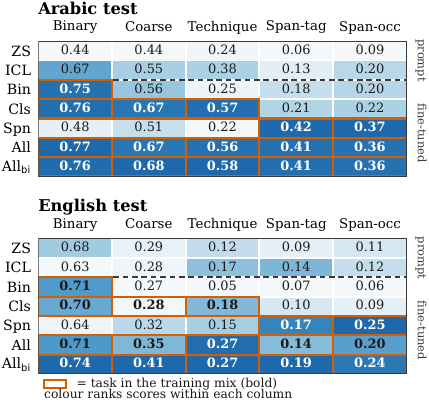}
\vspace{-0.2cm}
\caption{Supervision transfer with Qwen2.5-7B-Instruct across prompting and fine-tuning settings. Outlined cells mark trained tasks, and colors are normalized by column. Table~\ref{tab:transfer-full} reports the full results.}
% \caption{Supervision transfer, Qwen2.5-7B-Instruct. Rows: ZS zero-shot, ICL five same-language demonstrations, then fine-tuning on Bin binary, Cls the three classification tasks, Spn both span formats, All all five, All$_{\mathrm{bi}}$ all five in both languages. Outlined bold cells are tasks the model was trained for; colour is normalised per column. Absolute values in Table~\ref{tab:transfer-full}.}
\label{fig:rq2}
\vspace{-0.4cm}
\end{figure}

In Figure~\ref{fig:rq2}, we show that joint supervision generally outperforms single-task training. Classification tasks reinforce one another, preserving binary performance while improving coarse- and technique-level prediction. Span-only training weakens binary detection, whereas classification-only training fails to recover accurate span boundaries, indicating that sentence- and token-level supervision provide complementary signals. Training on all five tasks combines these benefits. \ourbest{} matches or exceeds task-specific models in most settings, with small losses on span prediction and the largest gains on English classification tasks. Overall, multi-task learning yields the strongest and most balanced performance across languages and tasks.

\subsection{Data Expansion \& Cross-Lingual Transfer}
% RQ3 covers data expansion and bilingual transfer.

\noindent\textbf{English data expansion improves performance.}
As shown in Table~\ref{tab:dataval}, expanding the English training set improves results on the new benchmark. Binary-only training overfits and degrades on the original test set, whereas multi-task training improves both benchmarks.

\begin{figure}[t!]
\centering
\includegraphics[width=0.8\columnwidth]{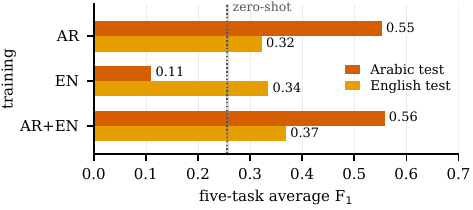}
\vspace{-0.2cm}
\caption{Five-task average by training language, Qwen2.5-7B-Instruct. English-only fine-tuning falls below the zero-shot level on Arabic.}
\label{fig:rq3}
\vspace{-0.4cm}
\end{figure}

\noindent\textbf{Cross-lingual transfer is asymmetric.}
Figure~\ref{fig:rq3} shows that transfer across languages is highly asymmetric. Arabic fine-tuning transfers effectively to English, especially for structured prediction tasks, while English-only fine-tuning transfers poorly to Arabic and substantially degrades technique and span prediction.

\noindent\textbf{Joint bilingual training is most robust.}
Joint training achieves the strongest overall performance across both languages and avoids the severe Arabic degradation caused by English-only fine-tuning. It therefore provides the best balance between task performance and cross-lingual generalization.

\begin{table}[!tbh]
\centering
\small
\setlength{\tabcolsep}{4.5pt}
\scalebox{0.8}{
\begin{tabular}{@{}lcc@{}}
\toprule
 & \textbf{\dsname{}} & \textbf{Original} \\
\textbf{English training data} & \textbf{EN test} & \textbf{EN test} \\
\midrule
Original, binary only          & 0.582 & 0.641 \\
Expanded, binary only          & 0.711 & 0.593 \\
Expanded, multi-task bilingual & \textbf{0.735} & \textbf{0.685} \\
\bottomrule
\end{tabular}
}
\vspace{-0.2cm}
\caption{English binary macro-F$_1$ under a fixed training recipe. Data expansion improves \dsname{}, while only multi-task training generalizes to the original PropXplain test set.}
% \caption{English binary macro-F$_1$ under one fixed recipe. The expansion helps on \dsname, and only multi-task training also holds on the original test set of \citet{hasanain2025propxplain}.}
\label{tab:dataval}
\vspace{-0.3cm}
\end{table}

% \subsection{Cross-Level Consistency and Error Analysis}
% % \todo{add occ and json figure and analysis}
% Additional analysis supporting RQ1 and RQ2. Appendix~\ref{app:explanations} evaluates the generated explanations against the references, and Appendix~\ref{app:longtail} analyzes the long-tailed technique distribution behind the fine-grained scores.

\subsection{Distillation and Direct Supervision}
\label{sec:distill}

We test whether distillation can replace direct task supervision and we evaluate two approaches. \textit{(i)} \textit{Off-policy distillation} trains Qwen2.5-7B to imitate GPT-5 reasoning traces generated from the reference annotations \cite{hsieh2023distilling}. \textit{(ii)} \textit{On-policy self-distillation (OPSD)} instead matches a student conditioned on the input to a teacher conditioned on both the input and gold annotation \cite{zhao2026opsd}, as illustrated in Figure~\ref{fig:opsd}. Following \cite{zhao2026opsd}, OPSD clips each token-level forward-KL contribution

% \begin{equation}
% \mathcal{L}
% =
% \frac{1}{|\hat{y}|}
% \sum_{n=1}^{|\hat{y}|}
% \sum_{v\in V}
% \min(\ell_{n,v},\tau),
% \label{eq:opsd}
% \end{equation}
% where \(\ell_{n,v}=p_T(v)[\log p_T(v)-\log p_S(v)]\) and the original method sets \(\tau=0.06\).
\begingroup
\footnotesize
\begin{equation}
\mathcal{L}
=
\frac{1}{|\hat{y}|}
\sum_{n=1}^{|\hat{y}|}
\sum_{v\in V}
\min(\ell_{n,v},\tau),
\label{eq:opsd}
\end{equation}
\endgroup
where \(\ell_{n,v}=p_T(v)[\log p_T(v)-\log p_S(v)]\), and the released configuration sets \(\tau=0.06\).

\begin{table}[!tbh]
\centering
\small
\setlength{\tabcolsep}{3.0pt}
\scalebox{0.8}{
\begin{tabular}{@{}llcc@{}}
\toprule
\textbf{Starting point} & \textbf{Training} & \textbf{Bin} & \textbf{Avg.} \\
\midrule
\multirow{3}{*}{Qwen2.5-7B}
 & none (zero-shot)                 & 0.561 & 0.256 \\
 & off-policy, GPT-5 traces         & 0.697 & 0.441 \\
 & OPSD, best configuration         & 0.623 & 0.288 \\
\midrule
\multirow{4}{*}{Qwen3-8B}
 & none (zero-shot)                 & 0.490 & 0.295 \\
 & OPSD, hard \(\tau=0.06\)         & 0.499 & 0.298 \\
 & OPSD, hard \(\tau=0.5\)          & 0.485 & 0.295 \\
 & OPSD, soft \(\tau=0.5\)        & \textbf{0.588} & \textbf{0.314} \\
\midrule
\multirow{3}{*}{\ourbest}
 & none                             & 0.749 & 0.464 \\
 & OPSD, hard \(\tau=0.5\)          & 0.742 & 0.429 \\
 & OPSD, soft \(\tau=0.5\)        & 0.751 & 0.464 \\
\bottomrule
\end{tabular}
}
\vspace{-0.3cm}
\caption{Distillation results across starting points. Bin denotes binary macro-F$_1$, and Avg.\ the five-task average across languages. Appendix~\ref{app:distill} reports full results.}
% \caption{Distillation results relative to each starting point. Bin denotes binary macro-F$_1$, and Avg.\ denotes the five-task average across Arabic and English. Bold marks the gain confirmed through seed replication. Appendix~\ref{app:distill} reports the full results.}
\label{tab:distill}
\vspace{-0.3cm}
\end{table}

\noindent\textbf{Direct supervision remains strongest.}
In Table~\ref{tab:distill}, off-policy distillation substantially improves the zero-shot model, whereas OPSD provides limited gains. Neither approach matches \ourbest{}, showing that distillation complements rather than replaces direct supervision.

\noindent\textbf{Hard clipping suppresses decision-token gradients.}
In Figure~\ref{fig:opsd}, we find that teacher--student disagreement concentrates on a small number of decision tokens with KL contributions above the clipping threshold. Hard clipping removes their gradients, while increasing \(\tau\) adds limited useful supervision.

\noindent\textbf{Saturating clipping improves OPSD.}
We replace hard clipping with
% \begin{equation}
% f_\tau(\ell)
% =
% \tau\log\!\left(1+\frac{\ell}{\tau}\right),
% \label{eq:soft}
% \end{equation}
\begingroup
\footnotesize
\begin{equation}
f_\tau(\ell)
=
\tau\log\!\left(1+\frac{\ell}{\tau}\right),
\label{eq:soft}
\end{equation}
\endgroup
which reduces large contributions without eliminating their gradients. 
% This formulation improves Qwen3-8B and prevents OPSD from degrading \ourbest{}, although it still falls short of direct supervision. Appendix~\ref{app:distill} provides the full configuration grid and results on HateXplain and GSM8K.
This consistently improves OPSD, recovering Qwen3-8B's English binary and increasing the average performance, with similar trends on HateXplain \cite{mathew2021hatexplain} and GSM8K \cite{cobbe2021gsm8k}. Nevertheless, even with the proposed clipping function, distillation remains complementary to rather than a replacement for direct supervision. Additional analyses are provided in Appendix~\ref{app:distill}.

\begin{table}[!tbh]
\centering
\small
\setlength{\tabcolsep}{3.5pt}
\scalebox{0.8}{
\begin{tabular}{@{}lcccc@{}}
\toprule
\textbf{Testbed (metric)} & $\boldsymbol{\tau}$ & \textbf{Zero-shot} & \textbf{Hard} & \textbf{Soft} \\
\midrule
\dsname{} EN binary (macro-F$_1$) & 0.5  & 0.449 & 0.448 & \textbf{0.692} \\
HateXplain (macro-F$_1$)          & 0.5  & 0.604 & 0.666 & \textbf{0.687} \\
GSM8K (accuracy)                  & 0.06 & 0.918 & 0.900 & \textbf{0.928} \\
AIME/HMMT (Avg@12)                & 0.05 & 34.2  & \textbf{41.9} & 39.8 \\
\bottomrule
\end{tabular}
}
\caption{Hard versus saturating (soft) clipping on four testbeds. Zero-shot is the model before OPSD training; each row uses its own metric and threshold $\tau$ (the published value on GSM8K and the original AIME/HMMT benchmarks, $\tau{=}0.5$ elsewhere). Best per row in bold. Appendix~\ref{app:distill} reports the full grids.}
\label{tab:clipgrid}
\vspace{-0.3cm}
\end{table}

\section{Comparison with Published Results}
\label{app:sota}

In Table~\ref{tab:sota}, we compare our bilingual multi-task model with the PropXplain systems that, like ours, predict the label \emph{and} generate an explanation \cite{hasanain2025propxplain}, on identical test splits. \ourbest{} obtains the best macro-F$_1$ in both languages while additionally producing techniques and spans from the same checkpoint. 

Table~\ref{tab:sota-span} compares span detection against the published GPT-4 results of \citet{hasanain-etal-2024-gpt}. The Arabic comparison is on the identical test split. For English no published result exists on our test set, which is introduced in this work; the closest published number is on the SemEval-2023 development set, and is marked as a different split. SemEval-2020 span identification omits technique labels, and SemEval-2023 techniques \cite{piskorski2023semeval,lepekhin2023ftd} are paragraph-level, so neither is directly comparable.

\begin{table}[htbp]
\centering
\small
\setlength{\tabcolsep}{4pt}
\scalebox{0.8}{
\begin{tabular}{@{}lcc@{}}
\toprule
\textbf{System} & \textbf{Macro-F$_1$} & \textbf{F$_1$\textsubscript{BERT}$^{\ddagger}$} \\
\midrule
\multicolumn{3}{@{}l}{\emph{\textbf{Arabic}, same test split}} \\
Llama-3.1-8B (base) \cite{hasanain2025propxplain} & 0.588 & 0.507 \\
Llama-3.1-8B (FT) \cite{hasanain2025propxplain}   & 0.760 & 0.706 \\
\rowcolor{gray!15}
\ourbest{} (ours; five tasks, two languages)       & \textbf{0.763} & 0.664 \\
\midrule
\multicolumn{3}{@{}l}{\emph{\textbf{English}, original PropXplain test set}} \\
Llama-3.1-8B (base) \cite{hasanain2025propxplain} & 0.562 & 0.596 \\
Llama-3.1-8B (FT) \cite{hasanain2025propxplain}   & 0.675 & 0.751 \\
\rowcolor{gray!15}
\ourbest{} (ours; five tasks, two languages)       & \textbf{0.685} & 0.718 \\
\bottomrule
\end{tabular}
}
\vspace{-0.2cm}
\caption{Comparison with PropXplain on matched test splits. Bold marks the best macro-F$_1$. $^{\ddagger}$Explanation BERTScores use different backbones across studies and are not directly comparable.}
% \caption{Label-and-explanation systems from PropXplain versus ours on identical test splits (best macro-F$_1$ in bold). $^{\ddagger}$Explanation BERTScores are computed with different backbones across the two papers and are not on one scale.}

\label{tab:sota}
\vspace{-0.3cm}
\end{table}

\begin{table}[htbp]
\centering
\small
\setlength{\tabcolsep}{5pt}
\scalebox{0.8}{
\begin{tabular}{@{}lcc@{}}
\toprule
\textbf{System} & \textbf{AR} & \textbf{EN} \\
\midrule
GPT-4, zero-shot \cite{hasanain-etal-2024-gpt} & 0.117 & 0.111$^{*}$ \\
\rowcolor{gray!10}
Task-specific SFT (ours)                  & \textbf{0.421} & \textbf{0.255} \\
\rowcolor{gray!10}
\ourbest{} (ours)                          & 0.411 & 0.241 \\
\bottomrule
\end{tabular}
}
\vspace{-0.2cm}
\caption{Technique-labeled span detection using overlap-adjusted micro-F$_1$. Arabic uses the same test split. $^{*}$The English GPT-4 score uses the SemEval-2023 development set and is not directly comparable.}
% \caption{Technique-labeled span detection, overlap-adjusted micro-F$_1$ (best of the two output formats per system). Arabic: identical test split. $^{*}$The English GPT-4 result is on the SemEval-2023 development set, a different split; no published results exist on our English span test set, which is introduced in this work.}
\label{tab:sota-span}
\vspace{-0.3cm}
\end{table}

\section{Conclusions and Future Work}
\label{sec:conclusions}

We frame propaganda analysis around four interrelated questions. Is propaganda present, which techniques appear, where do they occur, and why do the identified spans support the prediction? To study these questions jointly, we introduce \dsname, an Arabic--English resource that extends PropXplain with aligned binary, multi-label, span-level, and explanation annotations. Our experiments show that bilingual multi-task training provides the strongest and most balanced performance across tasks and languages, while simplifying deployment through a single model.
% We will release the dataset, code, and evaluation scripts to support reproducible research. 
Future work will extend the resource to additional languages and domains, improve span and explanation generation, and investigate more effective transfer and distillation methods for fine-grained propaganda analysis.

\section*{Limitations}
Our study focuses on Arabic and English propaganda in news sentences and social media posts. We use a shared taxonomy of 23 techniques to enable controlled comparisons across tasks and languages. As span annotation requires careful judgment, particularly at ambiguous boundaries, we use multiple annotators and expert review to improve consistency. Reference explanations provide valuable supervision and evaluation targets. Our experiments cover several representative model families and training settings. Future work can extend this analysis to additional architectures, domains, languages, and methods for evaluating explanation faithfulness.

\section*{Ethics and Broader Impact}
Propaganda detection can support media analysis, fact-checking, and research on harmful or manipulative communication. The same systems may also produce incorrect labels or be used to suppress legitimate criticism, satire, or political expression. We therefore recommend using these models to support trained human reviewers, particularly in high-stakes moderation and policy settings.
The dataset may contain political, sensitive, or offensive content from news and social media sources. Users should handle the data carefully and consider the cultural and political context of each instance. We will release data, code, and evaluation scripts to improve transparency and reproducibility. We also encourage users to report results separately across languages and tasks to avoid masking uneven performance.

% Models trained on this dataset could serve as valuable tools for fact-checkers, journalists, and social media platforms.

\section*{Acknowledgments}
The work of F. Alam, M. Hasanain, and F. Ahmed is supported by the NPRP grant 14C-0916-210015 from the Qatar National Research Fund part of Qatar Research Development and Innovation Council (QRDI). The findings achieved herein are solely the responsibility of the authors.

\bibliography{bibliography/main,bibliography/references}

\appendix
\section{Comparison with Prior Work}
\label{app:prior_work}
In Table~\ref{tab:resource-comparison}, we compare the task coverage of closely related propaganda resources. Prior datasets typically support only a subset of binary classification, technique classification, span identification, and natural language explanation generation. In contrast, \dsname combines Arabic and English data with binary labels, a shared taxonomy of 23 techniques, technique-annotated spans, and reference explanations.

\begin{table}[!tbh]
\centering
\scriptsize
\setlength{\tabcolsep}{2.0pt}
\scalebox{0.9}{
\begin{tabular}{@{}p{5.0cm}ccccc@{}}
\toprule
\textbf{Resource} & \textbf{Lang.} & \textbf{Bin.} & \textbf{Tech.} & \textbf{Span} & \textbf{NLE} \\
\midrule
Da San Martino et al. \cite{DaSanMartino2019emnlp} & EN & \checkmark & 18 & \checkmark & -- \\
SemEval-2020 \cite{DaSanMartino2020semeval} & EN & -- & 14 & \checkmark & -- \\
SemEval-2023 \cite{piskorski2023semeval} & 9L & -- & 23 & -- & -- \\
ArPro \cite{hasanain-etal-2024-gpt} & AR & \checkmark & 23 & \checkmark & -- \\
SemEval-2024 \cite{dimitrov2024semeval} & 4L & \checkmark & 22 & -- & -- \\
PropXplain \cite{hasanain2025propxplain} & AR/EN & \checkmark & -- & -- & \checkmark \\
\midrule
\textbf{\dsname} & AR/EN & \checkmark & 23 & \checkmark & \checkmark \\
\bottomrule
\end{tabular}
}
\vspace{-0.2cm}
\caption{Task coverage in prior propaganda resources. \textbf{Bin.:} binary sentence- or paragraph-level classification; \textbf{Tech.:} the number of fine-grained techniques; \textbf{Span:} technique-annotated text spans; \textbf{NLE:} natural language explanations. 9L/4L: nine/four languages.}
\label{tab:resource-comparison}
\vspace{-0.2cm}
\end{table}

\section{\dsname{} Additional Details}
\label{sec_appx_data_additional_details}

Table~\ref{tab:annotation-statistics} summarizes the annotation statistics by language. Arabic instances are longer on average and contain more techniques and annotated spans than English instances. They also show substantially higher proportions of multi-technique and multi-span annotations, indicating greater annotation complexity.

\begin{table}[t]
\centering
% \scriptsize
\setlength{\tabcolsep}{2.2pt}
\scalebox{0.8}{
\begin{tabular}{@{}lrrrrr@{}}
\toprule
\textbf{Lang.} & \textbf{Text} & \textbf{Tech.} & \textbf{Span} & \textbf{Multi-T} & \textbf{Multi-S} \\
\midrule
AR & 32.6 & 1.8 & 3.0 & 53.3 & 70.7 \\
EN & 23.8 & 1.3 & 1.5 & 24.9 & 30.6 \\
\bottomrule
\end{tabular}
}
\vspace{-0.2cm}
\caption{Annotation statistics by language. \textbf{Text:} the mean input length in words. \textbf{Tech.} and \textbf{Span} represent mean counts over propagandistic instances. \textbf{Multi-T} and \textbf{Multi-S} represent the percentages of propagandistic instances with more than one technique or span.}
\label{tab:annotation-statistics}
\vspace{-0.2cm}
\end{table}

\begin{figure}[!tbh]
\centering
\includegraphics[width=0.9\columnwidth]{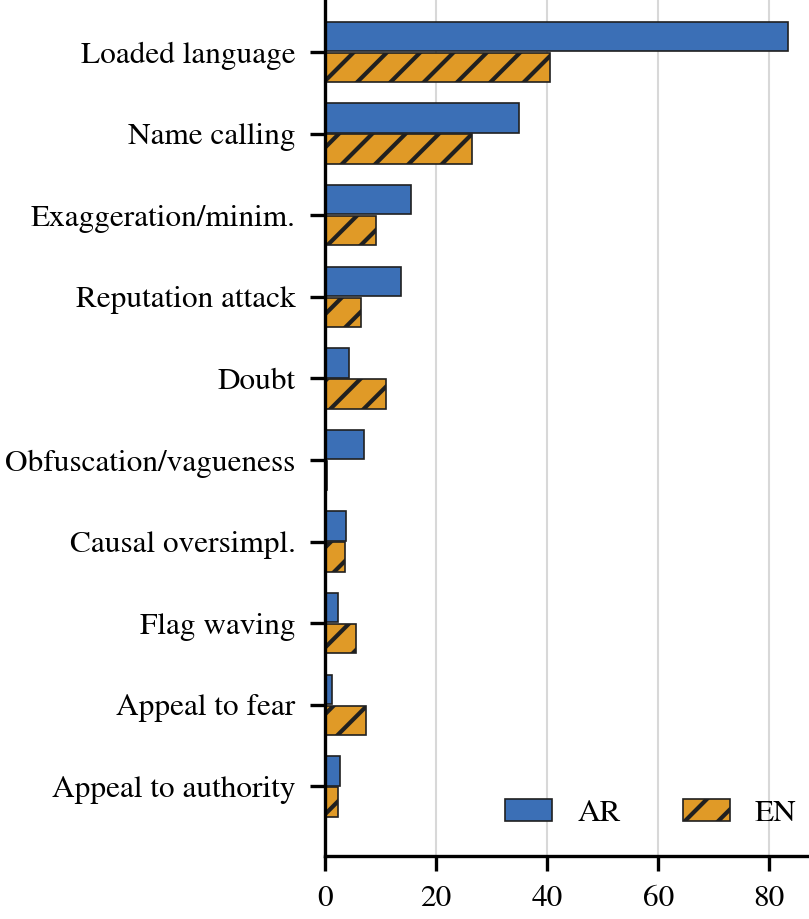}
\vspace{-0.2cm}
\caption{Frequency of the ten most common techniques among propagandistic instances.
% Percentages may sum above 100\% because each instance can contain multiple techniques.
}
\label{fig:technique-distribution}
\vspace{-0.3cm}
\end{figure}

\section{Label Inventory and Task Derivation}
\label{app:labels}

\begin{table}[htbp]
\centering
\small
\setlength{\tabcolsep}{3pt}
\renewcommand{\arraystretch}{1.05}
\begin{tabular}{@{}lp{0.55\columnwidth}@{}}
\toprule
Coarse category & Fine-grained techniques \\
\midrule
Manipulative Wording & Loaded Language, Exaggeration--Minimisation, Obfuscation--Vagueness--Confusion, Repetition \\
Reputation & Name Calling--Labeling, Questioning the Reputation, Doubt, Guilt by Association, Appeal to Hypocrisy \\
Justification & Appeal to Authority, Appeal to Fear--Prejudice, Appeal to Popularity, Appeal to Values, Flag Waving \\
Simplification & Causal Oversimplification, Consequential Oversimplification, False Dilemma--No Choice \\
Call & Slogans, Conversation Killer, Appeal to Time \\
Distraction & Red Herring, Straw Man, Whataboutism \\
\bottomrule
\end{tabular}
\caption{The shared label inventory: 23 techniques in six coarse categories, identical for both languages.}
\label{tab:labelmap}
\end{table}

Both languages share a unified inventory of 23 propaganda techniques grouped into six coarse categories (Table~\ref{tab:labelmap}), following the SemEval-2023 persuasion taxonomy \cite{piskorski2023semeval}, which was adopted for Arabic by ArPro \cite{hasanain-etal-2024-gpt}. Per-technique test support is given in Table~\ref{tab:pertech}.

\paragraph{Hierarchical annotation.} Span annotations are the primary labels. A sentence is propagandistic if it contains at least one annotated span; fine- and coarse-grained labels are derived from the span techniques. We verified this hierarchy for every sentence in both languages, ensuring that all annotation levels are consistent by construction.

\paragraph{Annotation characteristics.} Arabic annotations are denser than English, with more spans, distinct techniques, and repeated techniques per sentence, as well as shorter spans. This increases structured prediction complexity and motivates the span-occ representation.

\section{Experimental Setup}
\label{app:setup}

Table~\ref{tab:hyper} lists every configuration used in the paper. We fix one recipe before running the ablations and reuse it for every fine-tuned model, so no comparison mixes recipes. All generative systems share Qwen2.5-7B-Instruct, emit one fixed textual format per task, and are scored by a single parser, so the differences we report come from supervision rather than from decoding or evaluation choices.

\begin{table*}[t]
\centering
\footnotesize
\setlength{\tabcolsep}{4pt}
\begin{minipage}[t]{0.48\textwidth}
\centering
\begin{tabular}[t]{@{}ll@{}}
\toprule
Parameter & Value \\
\midrule
\multicolumn{2}{@{}l}{\emph{Fine-tuning (identical for every regime)}} \\
Backbone & Qwen2.5-7B-Instruct \\
Adapter & LoRA $r{=}16$, $\alpha{=}32$ \\
Learning rate & $1\times10^{-5}$, warmup ratio 0.05 \\
Epochs / batch & 4 / 4 per device (4 GPUs) \\
Max sequence length & 4{,}096 \\
Checkpoint selection & minimum validation loss \\
\midrule
\multicolumn{2}{@{}l}{\emph{Training regimes}} \\
Binary only & binary \\
Classification & binary, coarse, technique \\
Spans only & span-tag, span-occ \\
All five & all tasks above \\
Languages & Arabic, English, and both \\
\midrule
\multicolumn{2}{@{}l}{\emph{In-context learning}} \\
Retrievers & random, BM25, LaBSE, \\
 & mE5, BGE-M3, hybrid \\
Pools & same-language, cross-lingual, \\
 & mixed \\
Demonstrations & $k\in\{1,3,5\}$, most similar last \\
Selection & 18 combinations, ranked \\
 & on dev at $k{=}3$ \\
\midrule
\multicolumn{2}{@{}l}{\emph{Discriminative baselines}} \\
Heads & Llama-3.1-8B, Qwen2.5-7B \\
 & + linear head \\
Encoders & AraBERT-v2 (AR), \\
 & BERT-base (EN) \\
\bottomrule
\end{tabular}
\end{minipage}\hfill
\begin{minipage}[t]{0.48\textwidth}
\centering
\begin{tabular}[t]{@{}ll@{}}
\toprule
Parameter & Value \\
\midrule
\multicolumn{2}{@{}l}{\emph{Off-policy distillation (Appendix~\ref{app:distill})}} \\
Teacher traces & GPT-5, rationalize gold \\
Recipe & as fine-tuning; 8 epochs; \\
 & dev-selected \\
\midrule
\multicolumn{2}{@{}l}{\emph{OPSD (original recipe unless noted)}} \\
Adapter & LoRA $r{=}64$, $\alpha{=}128$, all proj. \\
Learning rate & $5\times10^{-6}$; max gradient norm 0.1 \\
Effective batch & 32 (2/device, accum.\ 4, 4 GPUs) \\
Divergence & forward KL, full vocabulary \\
Rollouts & temp.\ 1.1, top-$p$ 0.95, top-$k$ 20, \\
 & budget 1{,}024 tokens \\
Steps & 150, ckpt.\ every 10 (ext.: 450) \\
Selection & development subset (360 items) \\
Clip & $\tau\in\{0,.06,.1,.2,.5,1\}$, \\
 & hard or saturating \\
Seeds & 42; replicates 13, 77 \\
 & on the pivotal contrast \\
\midrule
\multicolumn{2}{@{}l}{\emph{Evaluation}} \\
Decoding & greedy \\
Binary & macro-F$_1$ \\
Coarse, technique & micro-F$_1$ \\
Spans & overlap-adjusted micro-F$_1$ \\
Explanations & BERTScore-F$_1$, BLEU, METEOR \\
\bottomrule
\end{tabular}
\end{minipage}
\caption{Configurations used throughout the paper.}
\label{tab:hyper}
\end{table*}

\section{Span Representations: Details and Results}
\label{app:spanformats}

Span-tag preserves sentence context and is generally more robust, whereas span-occ naturally supports overlapping spans and repeated strings. The two formats exhibit complementary failure modes: span-tag can break alignment when the model edits the copied sentence, while span-occ fails when generated spans do not exactly match the input, particularly in Arabic, where spans are shorter and denser. Encoder baselines instead use token-level indexing, which is naturally aligned with BIO tagging; we did not use character-offset indexing with LLMs, whose tokenized view makes predicted offsets unreliable \cite{semin2026strategies}. Table~\ref{tab:spanres} compares all systems on the span task under both representations.

\begin{table}[htbp]
\centering
\small
\setlength{\tabcolsep}{2.4pt}
\scalebox{0.85}{
\begin{tabular}{@{}l ccc ccc@{}}
\toprule
 & \multicolumn{3}{c}{Arabic} & \multicolumn{3}{c}{English} \\
\cmidrule(lr){2-4}\cmidrule(l){5-7}
System & tag & occ & BIO & tag & occ & BIO \\
\midrule
Whole sentence as span   & 0.082 & 0.082 & 0.082 & 0.038 & 0.038 & 0.038 \\
AraBERT-v2 / BERT-base   & --   & --   & 0.249 & -- & --  & 0.227 \\
Qwen2.5-7B zero-shot     & 0.058 & 0.093 & -- & 0.087 & 0.108 & -- \\
Qwen2.5-7B ICL, $k{=}5$  & 0.128 & 0.200 & -- & 0.142 & 0.122 & -- \\
GPT-5 (CoT)              & --   & 0.218 & -- & --   & 0.180 & -- \\
Task-specific SFT        & \textbf{0.421} & \textbf{0.374} & -- & 0.173 & \textbf{0.255} & -- \\
\ourbest{}               & 0.411 & 0.362 & -- & \textbf{0.189} & 0.241 & -- \\
\bottomrule
\end{tabular}
}
\caption{Technique-labeled span detection (overlap-adjusted micro-F$_1$; best per column in bold). Tag and occ are the generative formats; BIO is the encoders' token-level tagging. GPT-5 emits one span list, scored under the occ protocol.}
\label{tab:spanres}
\end{table}

Both span formats are built from the same gold spans (Algorithm~\ref{alg:spanbuild}), parsed back to character offsets by one routine per format (Algorithm~\ref{alg:spanparse}), and scored by the same metric, so the representation is the only variable between them.

\begin{algorithm}[h]
\caption{Target construction}
\label{alg:spanbuild}
\begin{algorithmic}[1]
\Statex \textbf{Span-tag}
\State sort gold spans by start offset, descending
\For{each span $(s, e, \ell)$}
    \State insert \texttt{<span type="$\ell$">} at $s$, \texttt{</span>} at $e$
\EndFor
\Statex \emph{right-to-left insertion keeps offsets valid for disjoint spans;}
\Statex \emph{overlapping spans cannot be represented as well-formed inline}
\Statex \emph{markup and are carried only by span-occ}
\Statex \textbf{Span-occ}
\For{each span $(s, e, \ell)$}
    \State $w \gets$ sentence$[s{:}e]$;\quad $k \gets$ which occurrence of $w$ starts at $s$ (1st, 2nd, \dots)
    \State emit \texttt{\{"text": $w$, "label": $\ell$, "occurrence": $k$\}}
\EndFor
\end{algorithmic}
\end{algorithm}

\begin{algorithm}[h]
\caption{Output parsing}
\label{alg:spanparse}
\begin{algorithmic}[1]
\Statex \textbf{Span-tag}
\For{each \texttt{<span type="$\ell$">}$c$\texttt{</span>}, left to right}
    \State $\hat{s} \gets$ untagged output characters consumed so far
    \If{sentence$[\hat{s} : \hat{s}{+}|c|] = c$} accept $(\hat{s}, \hat{s}{+}|c|, \ell)$
    \Else{} accept the unused occurrence of $c$ closest to $\hat{s}$, if any
    \EndIf
\EndFor
\Statex \textbf{Span-occ}
\State recover the JSON array (strip fences, repair commas, close truncation)
\For{each object $(w, \ell, k)$}
    \State locate the $k$-th occurrence of $w$: exact match, then
    \Statex \hspace{\algorithmicindent}\hspace{\algorithmicindent} Arabic-normalized, then case-insensitive
\EndFor
\Statex \emph{both parsers drop unmatched, zero-length, and duplicate spans}
\end{algorithmic}
\end{algorithm}

\paragraph{Scoring.} Parsed predictions $S$ and gold spans $T$ are evaluated using the overlap-adjusted measure of \citet{DaSanMartino2020semeval}, where only label-matching pairs contribute their character overlap normalized by the span in focus.
\begin{align}
P &= \frac{1}{|S|}\sum_{s\in S}\sum_{t\in T}\frac{|s\cap t|}{|s|}\,\delta\!\left(\ell(s),\ell(t)\right),\notag\\
R &= \frac{1}{|T|}\sum_{s\in S}\sum_{t\in T}\frac{|s\cap t|}{|t|}\,\delta\!\left(\ell(s),\ell(t)\right),
\label{eq:oaf1}
\end{align}
with F$_1$ their harmonic mean; sums and span counts are aggregated over the whole test set (micro) after removing duplicate (start, end, label) triples.

\section{In-Context Learning Ablations}
\label{app:icl}

We select the in-context configuration in two stages so that the main comparison uses one justified setting rather than a tuned-per-cell best case.

\begin{figure}[htbp]
\centering
\includegraphics[width=0.94\columnwidth]{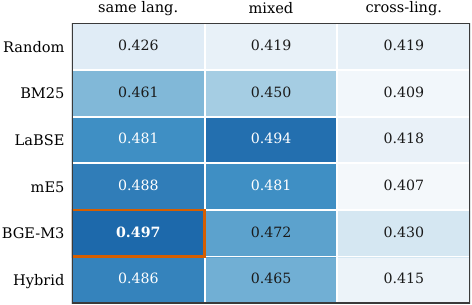}
\caption{Development ranking of retriever--pool combinations at $k{=}3$; the outlined cell is the winner.}
\label{fig:iclrank}
\end{figure}

\paragraph{Stage 1: which retriever and which pool.} We rank all 18 combinations of six retrievers and three demonstration pools on development data at $k{=}3$, scoring the mean of the primary metric over Arabic and English binary and technique classification (Figure~\ref{fig:iclrank}). Dense retrieval with BGE-M3 over a same-language pool ranks first (0.497). Every cross-lingual configuration scores within 0.004 of the random same-language baseline (0.426), with the best reaching only 0.430, and sparse retrieval with BM25 trails all dense retrievers.

\paragraph{Stage 2: how many demonstrations.} We run the winner, its cross-lingual counterpart, and a random control on the full test set at $k\in\{1,3,5\}$ (Table~\ref{tab:icl-full}). Retrieved same-language demonstrations improve monotonically with $k$ on Arabic for every task except binary, where $k{=}3$ is best. On English they do not help binary detection at any $k$. Cross-lingual demonstrations track the random baseline throughout, even though the English pool is 4.2 times larger than the Arabic one. When a same-language pool exists, even a small one, it should be used.

\begin{table}[htbp]
\centering
\small
\setlength{\tabcolsep}{3pt}
\begin{tabular}{@{}lccccc@{}}
\toprule
Configuration & Bin & Coa & Tech & S-tag & S-occ \\
\midrule
\multicolumn{6}{@{}l}{\emph{Arabic}} \\
Random, $k{=}5$        & 0.598 & 0.483 & 0.265 & 0.065 & 0.162 \\
BGE-M3 cross, $k{=}5$  & 0.622 & 0.501 & 0.268 & 0.051 & 0.107 \\
BGE-M3 same, $k{=}1$   & 0.613 & 0.512 & 0.286 & 0.085 & 0.170 \\
BGE-M3 same, $k{=}3$   & \textbf{0.688} & 0.536 & 0.349 & 0.114 & 0.190 \\
BGE-M3 same, $k{=}5$   & 0.669 & \textbf{0.550} & \textbf{0.376} & \textbf{0.128} & \textbf{0.200} \\
\midrule
\multicolumn{6}{@{}l}{\emph{English}} \\
Random, $k{=}5$        & 0.682 & 0.262 & 0.134 & 0.094 & 0.105 \\
BGE-M3 cross, $k{=}5$  & 0.628 & 0.238 & 0.140 & 0.115 & 0.106 \\
BGE-M3 same, $k{=}1$   & \textbf{0.695} & \textbf{0.286} & 0.151 & 0.109 & 0.105 \\
BGE-M3 same, $k{=}3$   & 0.655 & 0.276 & 0.158 & 0.125 & 0.119 \\
BGE-M3 same, $k{=}5$   & 0.632 & 0.281 & \textbf{0.172} & \textbf{0.142} & \textbf{0.122} \\
\bottomrule
\end{tabular}
\caption{In-context learning on the full test sets; \emph{same}/\emph{cross} is the demonstration-pool language.}
\label{tab:icl-full}
\end{table}

\section{Full Transfer Results}
\label{app:transfer}

Table~\ref{tab:transfer-full} gives the absolute values behind Figure~\ref{fig:rq2} for both test languages, including every cross-lingual cell: rows named by one language are fine-tuned on that language only, so their columns under the other language measure cross-lingual transfer.

\begin{table*}[t]
\centering
\small
\setlength{\tabcolsep}{4.2pt}
\begin{tabular}{@{}l ccccc ccccc@{}}
\toprule
 & \multicolumn{5}{c}{Arabic test} & \multicolumn{5}{c}{English test} \\
\cmidrule(lr){2-6}\cmidrule(l){7-11}
Training & Bin & Coa & Tech & S-tag & S-occ & Bin & Coa & Tech & S-tag & S-occ \\
\midrule
none (zero-shot)       & 0.439 & 0.437 & 0.238 & 0.058 & 0.093 & 0.683 & 0.290 & 0.125 & 0.087 & 0.108 \\
\midrule
AR, all five tasks     & \textbf{0.767} & 0.666 & 0.560 & 0.408 & 0.364 & 0.607 & 0.361 & \textbf{0.272} & 0.198 & 0.177 \\
EN, all five tasks     & 0.400 & 0.061 & 0.047 & 0.004 & 0.041 & 0.712 & 0.355 & 0.268 & 0.139 & 0.203 \\
AR+EN, all five tasks  & 0.763 & \textbf{0.682} & \textbf{0.575} & 0.411 & 0.362 & \textbf{0.735} & \textbf{0.410} & \textbf{0.272} & 0.189 & 0.241 \\
\midrule
AR, binary only        & 0.754 & 0.564 & 0.254 & 0.177 & 0.203 & 0.638 & 0.259 & 0.084 & 0.108 & 0.106 \\
EN, binary only        & 0.544 & 0.514 & 0.044 & 0.043 & 0.059 & 0.711 & 0.272 & 0.051 & 0.069 & 0.059 \\
AR, classification (3) & 0.759 & 0.667 & 0.566 & 0.207 & 0.216 & 0.608 & 0.350 & 0.252 & 0.110 & 0.112 \\
EN, classification (3) & 0.500 & 0.097 & 0.037 & 0.053 & 0.055 & 0.702 & 0.276 & 0.182 & 0.097 & 0.090 \\
AR, spans only         & 0.485 & 0.515 & 0.222 & \textbf{0.421} & \textbf{0.374} & 0.649 & 0.366 & 0.121 & \textbf{0.204} & 0.174 \\
EN, spans only         & 0.313 & 0.112 & 0.095 & 0.019 & 0.066 & 0.639 & 0.323 & 0.153 & 0.173 & \textbf{0.255} \\
\bottomrule
\end{tabular}
\caption{Every training regime evaluated on every task in both languages; best per column in bold.}
\label{tab:transfer-full}
\end{table*}

\section{Complete Baseline Results}
\label{app:baselines}

Table~\ref{tab:baselines-full} extends the binary comparison of \S\ref{sec:baselines} to all five tasks. Three notes. First, the proprietary models answer one multi-task prompt that produces all five outputs at once, while the open models answer one prompt per task; the two Qwen2.5-7B multi-task rows quantify this protocol difference, lifting its Arabic binary score from 0.439 to 0.642, within 0.004 of GPT-5 under the identical prompt. Second, the majority baseline exceeds every training-free system on Arabic coarse and technique classification (0.558 and 0.466), because one category and one technique dominate the Arabic label distribution; scores on these two tasks should be read against that floor rather than against zero. Third, classification heads stay level with \ourbest{} on Arabic and lead on English technique classification, and the encoders stay competitive on Arabic classification (0.654 coarse) while dropping sharply on English; both families produce no explanations, and neither produces occurrence-indexed spans. Random baselines average five seeds; the single span list of the multi-task prompt is scored under the occurrence protocol (hence span-tag~--), and encoder spans come from token-level tagging under the same overlap-adjusted metric. We include previously reported results only when they use directly comparable data splits and evaluation metrics.

\begin{table*}[t]
\centering
\small
\setlength{\tabcolsep}{3.6pt}
\begin{tabular}{@{}l ccccc ccccc@{}}
\toprule
 & \multicolumn{5}{c}{Arabic test} & \multicolumn{5}{c}{English test} \\
\cmidrule(lr){2-6}\cmidrule(l){7-11}
System & Bin & Coa & Tech & S-tag & S-occ & Bin & Coa & Tech & S-tag & S-occ \\
\midrule
\multicolumn{11}{@{}l}{\emph{No learning}} \\
Majority label         & 0.380 & 0.558 & 0.466 & 0.000 & 0.000 & 0.419 & 0.196 & 0.162 & 0.000 & 0.000 \\
Random, fair coin      & 0.499 & 0.244 & 0.093 & 0.000 & 0.000 & 0.470 & 0.101 & 0.031 & 0.000 & 0.000 \\
Whole sentence as span & --   & --   & --   & 0.082 & 0.082 & --   & --   & --   & 0.038 & 0.038 \\
\midrule
\multicolumn{11}{@{}l}{\emph{Zero-shot, open weights (one prompt per task)}} \\
Fanar-2-27B            & 0.409 & 0.513 & 0.320 & 0.081 & 0.103 & 0.431 & 0.290 & 0.203 & 0.116 & 0.121 \\
Llama-3.1-8B           & 0.672 & 0.482 & 0.274 & 0.103 & 0.097 & 0.437 & 0.218 & 0.098 & 0.074 & 0.080 \\
Qwen2.5-7B             & 0.439 & 0.437 & 0.238 & 0.058 & 0.093 & 0.683 & 0.290 & 0.125 & 0.087 & 0.108 \\
Qwen3-VL-8B            & 0.433 & 0.448 & 0.196 & 0.021 & 0.029 & 0.669 & 0.281 & 0.149 & 0.046 & 0.075 \\
Qwen3-VL-8B (think)    & 0.577 & 0.464 & 0.239 & 0.138 & 0.137 & 0.625 & 0.278 & 0.166 & 0.156 & 0.145 \\
Gemma-4-E4B            & 0.442 & 0.391 & 0.343 & 0.080 & 0.117 & 0.671 & 0.280 & 0.187 & 0.138 & 0.150 \\
\midrule
\multicolumn{11}{@{}l}{\emph{Zero-shot, proprietary (one multi-task prompt, single span list)}} \\
GPT-5                  & 0.595 & 0.420 & 0.309 & -- & 0.184 & 0.663 & 0.374 & 0.276 & -- & 0.196 \\
GPT-5 (CoT)            & 0.646 & 0.467 & 0.358 & -- & 0.218 & 0.657 & 0.356 & 0.252 & -- & 0.180 \\
Gemini-3.1-Pro         & 0.553 & 0.392 & 0.297 & -- & 0.200 & 0.661 & 0.369 & 0.284 & -- & 0.218 \\
Gemini-3.1-Pro (CoT)   & 0.505 & 0.305 & 0.218 & -- & 0.148 & 0.673 & 0.364 & 0.282 & -- & 0.213 \\
\midrule
\multicolumn{11}{@{}l}{\emph{Protocol bridge: open model under the identical multi-task prompt}} \\
Qwen2.5-7B (multi-task)      & 0.516 & 0.237 & 0.103 & -- & 0.035 & 0.616 & 0.241 & 0.121 & -- & 0.072 \\
Qwen2.5-7B (multi-task, CoT) & 0.642 & 0.248 & 0.188 & -- & 0.058 & 0.598 & 0.218 & 0.150 & -- & 0.084 \\
\midrule
\multicolumn{11}{@{}l}{\emph{Fine-tuned discriminative (one model per task and language)}} \\
Llama-3.1-8B + head    & 0.778 & 0.683 & 0.582 & n/a & n/a & 0.732 & 0.449 & 0.392 & n/a & n/a \\
Qwen2.5-7B + head      & 0.765 & 0.690 & 0.581 & n/a & n/a & 0.741 & 0.420 & 0.353 & n/a & n/a \\
AraBERT-v2 / BERT-base & 0.746 & 0.654 & 0.529 & 0.249 & n/a & 0.719 & 0.402 & 0.293 & 0.227 & n/a \\
\midrule
\ourbest{} (ours)      & 0.763 & 0.682 & 0.575 & 0.411 & 0.362 & 0.735 & 0.410 & 0.272 & 0.189 & 0.241 \\
\bottomrule
\end{tabular}
\caption{All baselines on all five tasks (binary macro-F$_1$; coarse and technique micro-F$_1$; spans overlap-adjusted micro-F$_1$).}
\label{tab:baselines-full}
\end{table*}

\section{Explanation Quality}
\label{app:explanations}

We score generated explanations against the reference explanations of the binary cell (Table~\ref{tab:expl}), with AraBERT-v2 as the BERTScore backbone for Arabic and BERT-base for English. Raw BERTScore is therefore not comparable across languages: pairing each reference with a \emph{random} same-language reference already scores 0.58 on Arabic and 0.63 on English. The near-floor Arabic zero-shot score has a simple cause: the zero-shot model answers in English for every Arabic input (100\% of test items), and such cross-language pairs fall below even the random same-language floor. Retrieved demonstrations largely fix the output language (2\% remain in English) and fine-tuning removes the issue entirely. The fine-tuned row uses the model trained on each language's own data; the bilingual \ourbest{} matches it within 0.002 on every measure. Against the floors, the fine-tuned models clear their language's baseline by the same margin in both, $+0.09$ Arabic and $+0.09$ English, while the seemingly strong English zero-shot score is only $+0.03$ above chance pairing. Reference-based measures do not assess whether an explanation reflects the model's own decision process, which we leave to future work.

\begin{table}[htbp]
\centering
\small
\setlength{\tabcolsep}{2.4pt}
\scalebox{0.9}{
\begin{tabular}{@{}lcccccc@{}}
\toprule
 & \multicolumn{2}{c}{BERTScore} & \multicolumn{2}{c}{BLEU} & \multicolumn{2}{c}{METEOR} \\
\cmidrule(lr){2-3}\cmidrule(lr){4-5}\cmidrule(l){6-7}
System & AR & EN & AR & EN & AR & EN \\
\midrule
Zero-shot           & 0.381 & 0.657 & 0.002 & 0.035 & 0.024 & 0.212 \\
ICL, $k{=}5$        & 0.626 & 0.672 & 0.057 & 0.050 & 0.205 & 0.254 \\
SFT (same lang.) & \textbf{0.663} & \textbf{0.718} & \textbf{0.102} & \textbf{0.103} & \textbf{0.254} & \textbf{0.327} \\
\bottomrule
\end{tabular}
}
\caption{Explanation quality against the reference explanations.}
\label{tab:expl}
\end{table}

\section{Long-Tailed Techniques}
\label{app:longtail}

Technique performance follows the frequency of the label (Table~\ref{tab:pertech}). \ourbest{} handles the two most frequent techniques well in Arabic, degrades on mid-frequency ones, and predicts almost none of the rare tail: of the 14 Arabic techniques with fewer than 20 test occurrences, 11 score zero, as do 10 of the 11 such techniques in English. Improving the tail is the clearest remaining target for this benchmark.

\begin{table}[htbp]
\centering
\small
\setlength{\tabcolsep}{2.2pt}
\scalebox{0.96}{
\begin{tabular}{@{}lcccc@{}}
\toprule
 & \multicolumn{2}{c}{Arabic} & \multicolumn{2}{c}{English} \\
\cmidrule(lr){2-3}\cmidrule(l){4-5}
Technique & Sup. & F$_1$ & Sup. & F$_1$ \\
\midrule
Loaded Language            & 675 & 0.798 & 441 & 0.440 \\
Name Calling--Labeling     & 206 & 0.562 & 291 & 0.291 \\
Exaggeration--Minimisation & 159 & 0.272 & 106 & 0.053 \\
Questioning the Reputation &  99 & 0.521 &  73 & 0.115 \\
Causal Oversimplification  &  58 & 0.033 &  40 & 0.000 \\
Doubt                      &  46 & 0.080 & 119 & 0.135 \\
Appeal to Fear--Prejudice  &  19 & 0.080 &  84 & 0.121 \\
\midrule
Tail (sup.\ $<$ 20)      &  14 tech. & 0.036 & 11 tech. & 0.009 \\
\bottomrule
\end{tabular}
}
\caption{Per-technique F$_1$ of \ourbest{} with test support. The last row averages the techniques with fewer than 20 test occurrences.}
\label{tab:pertech}
\end{table}

\section{Distillation}
\label{app:distill}

This appendix details the two distillation routes of Section~\ref{sec:distill} and reports every trained configuration. In brief, off-policy teacher imitation loses to direct references; OPSD stalls because hard clipping removes gradients on decision tokens; saturating clipping repairs it on \dsname{}; and replication on the original benchmarks confirms that the published clip remains effective there.

\paragraph{Off-policy route.} We generate chain-of-thought traces for the full training set with GPT-5, conditioned on the gold annotations so the traces rationalize the correct answer \cite{hsieh2023distilling}, and fine-tune the student to imitate them (the API exposes text, not distributions). The best bilingual model reaches 0.727 Arabic and 0.668 English binary macro-F$_1$ (Table~\ref{tab:distill}), below the no-reasoning \ourbest{} at 0.763 and 0.735, and 0.441 against 0.505 on the four tasks both output formats support: a stronger teacher's reasoning, imitated off-policy, loses to plain references \cite{zhao2026opsd}.

\begin{figure}[t!]
\centering
\includegraphics[width=\columnwidth]{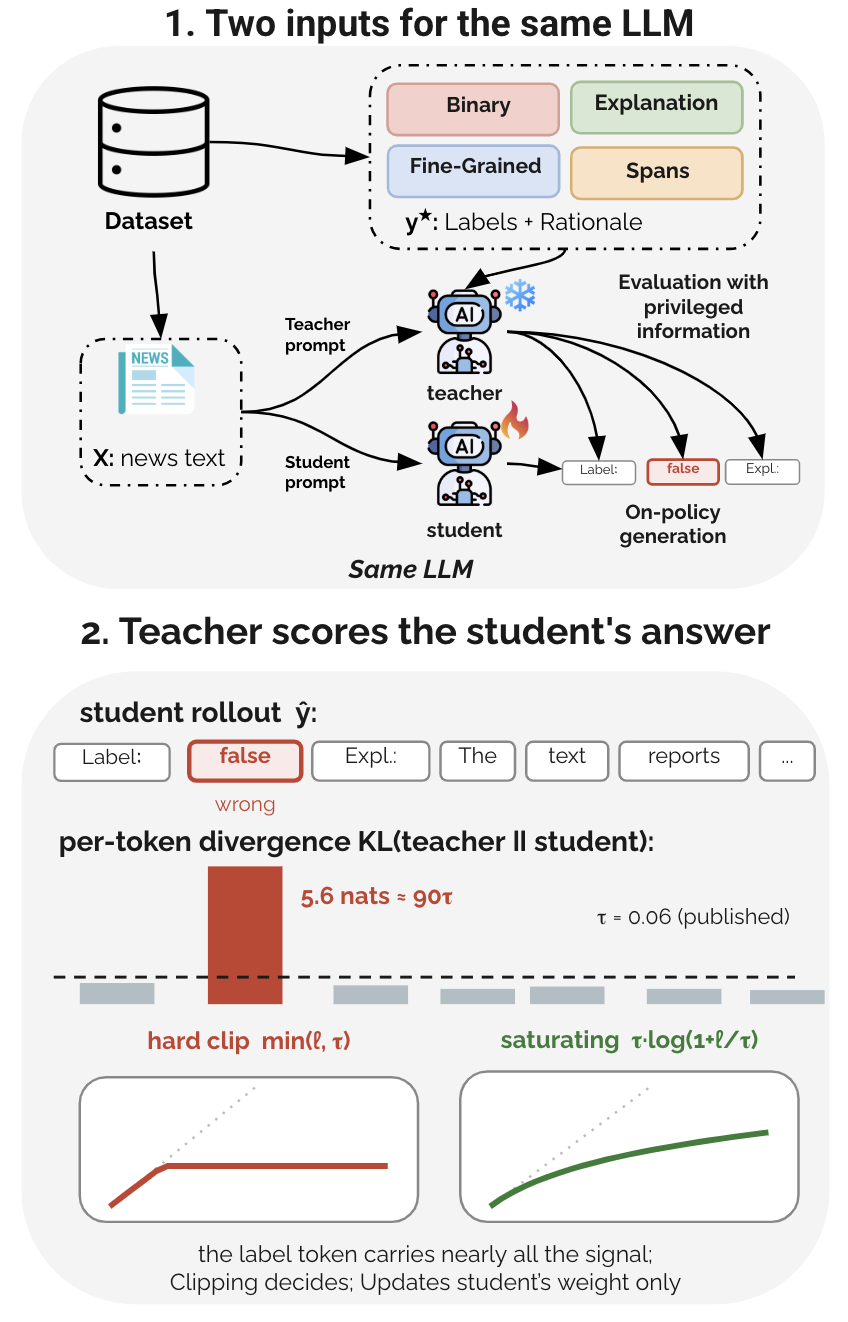}
\caption{Privileged self-distillation on \dsname{}. \textbf{Top:} one LLM in two roles: a student given only the news text $x$, and the same weights frozen with the gold annotation $y^{\star}$ in context. The teacher scores the student's own rollout, and only the student is updated. \textbf{Bottom:} disagreement concentrates on the wrong label token (the depicted token carries 5.6 nats, ${\approx}90\times$ the published clip $\tau=0.06$), exactly where the hard clip is flat and contributes zero gradient, while our saturating clip continues to provide a learning signal.}
\label{fig:opsd}
\vspace{-0.4cm}
\end{figure}

\paragraph{On-policy route (OPSD).} OPSD \cite{zhao2026opsd} uses one set of weights in two roles (Figure~\ref{fig:opsd}): the \emph{student} sees the task prompt $x$ and samples a rollout, while the \emph{teacher}, the same model frozen at initialization, additionally sees the verified gold block $g$ (labels, span quotes, and the reference explanation). Training pulls the student toward the teacher's better-informed next-token distributions via the clipped divergence of Equation~\ref{eq:opsd}, computed over the full vocabulary at every rollout position; gradients flow only through the student \cite{agarwal2024gkd}. The clip $\tau$ exists because stylistic tokens occasionally carry huge divergence; the released configuration fixes $\tau{=}0.06$, which the original work reports as untuned. We adopt the authors' released implementation and recipe (Table~\ref{tab:hyper}), changing only the data interface.

\paragraph{Why the hard clip can delete the signal.} The clip acts per vocabulary element, and $\partial\min(\ell,\tau)/\partial\ell = \mathbf{1}[\ell<\tau]$ almost everywhere: an element above threshold is not attenuated, it is removed from the gradient \cite{schulman2017ppo}. When the student emits a wrong label token, the teacher concentrates its mass on the single correct entry, and that one element carries nearly the whole per-token divergence: label-token divergence reaches 5.5 nats at the 90th percentile, about $90\times$ the published threshold. The saturating clip of Equation~\ref{eq:soft} (\emph{soft} in tables, versus \emph{hard} for $\min(\ell,\tau)$) grows only logarithmically on such outliers, but its gradient $f_\tau'(\ell) = 1/(1+\ell/\tau)$ never vanishes.

\paragraph{Reading the grid.} Table~\ref{tab:opsdgrid} lists all 21 trained OPSD configurations. For Qwen2.5 rows, think/no-think are prompt variants; for Qwen3 rows, sX/tY are the chat template's native thinking flags for student and teacher (default sOFF/tON); label-only privilege drops the reference explanation; LoRA rank is $r{=}64$ unless noted.

\paragraph{The clip's shape decides the outcome.} At the published settings OPSD is flat everywhere: the best of seven Qwen2.5-7B configurations gains $+0.032$ AVG over zero-shot, and it is one that \emph{removes} the clip; every configuration that keeps the published clip stays within $+0.001$. On \ourbest, the strongest initialization, the trained model agrees with its own starting point on 99.1\% of Arabic binary test predictions ($p{=}1.0$; Table~\ref{tab:significance}). The saturating clip is the only intervention that repairs the model consistently, and the effect replicates across three seeds: 0.314$\pm$0.003 AVG against 0.296$\pm$0.001 for the $\tau$-matched hard clip, driven by repairing Qwen3-8B's collapsed English binary head from 0.449 to 0.692$\pm$0.001 macro-F$_1$ ($p{<}0.001$), with no damage on \ourbest. The repair needs both the gradient-preserving shape and an adequate threshold: removing the clip entirely does not fix the English head (0.460), the saturating shape at the published $\tau{=}0.06$ does not either (0.445), the repair holds across $\tau\in\{0.1, 0.2, 0.5\}$, and at $\tau{=}1$, where the loss approaches the unclipped objective, it vanishes again (0.449). Across runs, the saturating clip repaired the head in all five runs ($\tau\in\{0.1,0.2,0.5\}$), whereas the hard clip left it collapsed in five of six; one $\tau{=}0.06$ seed partially escaped (0.668).

\begin{table}[htbp]
\centering
\small
\setlength{\tabcolsep}{2.6pt}
\begin{tabular}{@{}llcc@{}}
\toprule
Init & Configuration & Bin & AVG \\
\midrule
\multirow{7}{*}{\shortstack[l]{Qwen2.5\\(ZS 0.561/0.256)}}
 & think, hard $\tau{=}.06$    & 0.519 & 0.253 \\
 & think, no clip              & 0.514 & 0.245 \\
 & no-think, hard $\tau{=}.06$ & 0.574 & 0.257 \\
 & no-think, no clip           & 0.556 & 0.277 \\
 & no-think, no clip, $r{=}16$ & 0.623 & 0.288 \\
 & think, no clip, $r{=}16$    & 0.465 & 0.232 \\
 & label-only privilege        & 0.458 & 0.246 \\
\midrule
\multirow{10}{*}{\shortstack[l]{Qwen3-8B\\(ZS 0.490/0.295)}}
 & sOFF/tON, hard $\tau{=}.06$ & 0.499 & 0.298 \\
 & sON/tON, hard $\tau{=}.06$  & 0.521 & 0.302 \\
 & sOFF/tOFF, hard $\tau{=}.06$& 0.501 & 0.296 \\
 & hard $\tau{=}.06$, top-20 & 0.508 & 0.297 \\
 & sOFF/tON, hard $\tau{=}.5$  & 0.485 & 0.295 \\
 & sOFF/tON, soft $\tau{=}.1$  & 0.586 & 0.317 \\
 & sOFF/tON, soft $\tau{=}.2$  & 0.605 & 0.323 \\
 & sOFF/tON, soft $\tau{=}.5$  & 0.588 & 0.314 \\
 & sOFF/tON, soft $\tau{=}1$   & 0.494 & 0.296 \\
 & soft $\tau{=}.5$, 450 steps & 0.579 & 0.307 \\
\midrule
\multirow{4}{*}{\shortstack[l]{\ourbest\\(0.749/0.464)}}
 & hard $\tau{=}.06$           & 0.748 & 0.462 \\
 & hard $\tau{=}.5$            & 0.742 & 0.429 \\
 & soft $\tau{=}.5$            & 0.751 & 0.464 \\
 & soft $\tau{=}.5$, 450 steps & 0.741 & 0.449 \\
\bottomrule
\end{tabular}
\caption{All 21 trained OPSD configurations (mean of Arabic and English; each block's initialization in parentheses).}
\label{tab:opsdgrid}
\end{table}

\paragraph{The premise holds, and the signal lives in the label.} Could OPSD fail because the privileged context does not help the teacher? No. The gold block lifts label accuracy (exact match, 300 training-pool items over the three classification tasks) from 42.0\% to 93.7\% for the base Qwen2.5-7B and from 70.7\% to 94.3\% for \ourbest{} (Table~\ref{tab:oracle}). Label lines alone match or exceed the full block (98.3\% and 92.7\%), so the label itself is the carrier, and a \emph{mismatched} gold block collapses the base teacher to 20.3\%, repeating the asserted wrong label on 73\% of items. The teacher's distributions carry far more task information than the student's; whatever fails, it is not an absent signal.

\begin{table}[htbp]
\centering
\small
\setlength{\tabcolsep}{4.5pt}
\begin{tabular}{@{}lcc@{}}
\toprule
Teacher context & Qwen2.5-7B & \ourbest \\
\midrule
none (= student)          & 42.0 & 70.7 \\
full gold block           & 93.7 & 94.3 \\
\quad label lines only    & 98.3 & 92.7 \\
\quad explanation only    & 71.0 & 83.0 \\
\quad mismatched block    & 20.3 & 53.0 \\
\midrule
\emph{copies the mismatched label} & \emph{73.0} & \emph{38.7} \\
\bottomrule
\end{tabular}
\caption{Teacher label accuracy (\%) under different privileged contexts, on 300 training-pool items; the italic row is how often the teacher repeats a wrong asserted label.}
\label{tab:oracle}
\end{table}

\begin{figure*}[t]
\centering
\includegraphics[width=\textwidth]{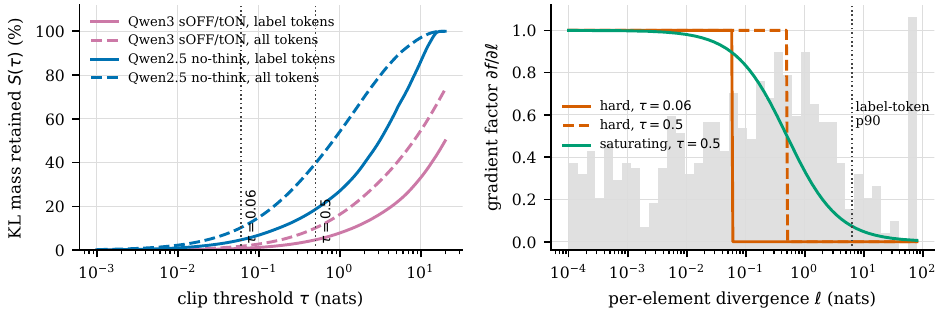}
\caption{Signal survival under the clip. \textbf{Left:} share of divergence a hard clip keeps, $S(\tau)$, for label tokens (solid) and all tokens (dashed). \textbf{Right:} each clip's gradient factor over the empirical label-token divergence (gray histogram); hard clips drop to zero at their thresholds. At the published $\tau{=}0.06$, only 1.8\% of the total divergence mass and 0.8\% of the label-token mass survive.}
\label{fig:mechanism}
\end{figure*}

\paragraph{The signal is tail-concentrated, and the clip deletes the tail.} We compute the per-position divergence on student rollouts and bucket tokens into \{label, quoted span, style, other\}. The median is tiny, 0.001 to 0.05 nats: the privileged teacher \emph{agrees} with the student almost everywhere, and the task signal concentrates in a thin tail on decision tokens, whose 90th-percentile divergence reaches 5.5 nats on Qwen3-8B. Figure~\ref{fig:mechanism} quantifies the consequence with the survival measure $S(\tau)=\sum\min(\ell^{+},\tau)/\sum\ell^{+}$: at $\tau{=}0.06$ the hard clip retains 1.8\% of this model's total divergence mass and 0.8\% of its label-token mass; even $\tau{=}0.5$ retains only 10.2\% and 4.7\%. Survival is below 1\% in \emph{every} content bucket, and the heaviest tails sit on the two most \emph{frequent} techniques, so no reweighting across techniques or classes can rescue the hard clip. The freeze is measurable end to end: the adapters reach a maximum entry of $4.4\times10^{-4}$ after 150 steps, development scores stay flat throughout training (checkpoint selection: $0.5{\times}$binary accuracy $+\,0.25{\times}$coarse $+\,0.25{\times}$fine-grained micro-F$_1$ on the held-out 360-item development subset), and trained students are 98.1 to 99.1\% prediction-identical to their initializations.

\begin{figure*}[t]
\centering
\includegraphics[width=\textwidth]{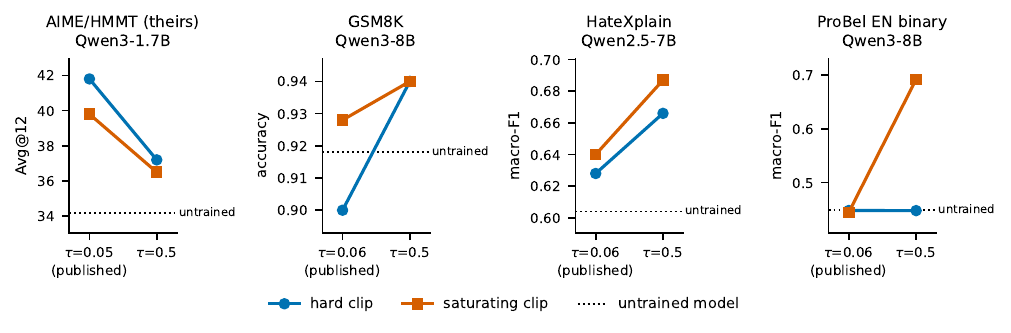}
\caption{Hard and saturating clips at the published $\tau$ (left) and $\tau{=}0.5$ (right); each panel uses its own metric, and dotted lines mark the untrained model. Points show medians when several seeds exist. On the original OPSD benchmark, hard clipping is better; on GSM8K, a larger threshold suffices for either shape; on HateXplain, saturating clipping is better at both thresholds; on \dsname{}, hard clipping remains near the untrained floor (one of three seeds at the published $\tau$ partially escaped; see text), while saturating clipping reaches 0.69 at $\tau{=}0.5$.}
\label{fig:generality}
\end{figure*}

\paragraph{The mechanism generalizes, and the clip choice follows the domain.} On HateXplain \cite{mathew2021hatexplain}, saturating clipping outperforms hard clipping at both thresholds (Figure~\ref{fig:generality}): 0.640 vs.\ 0.628 at the published $\tau$ and 0.687 vs.\ 0.666 at $\tau{=}0.5$ (untrained: 0.604). On GSM8K \cite{cobbe2021gsm8k}, the published hard clip degrades accuracy (0.918 to 0.900), while saturating clipping reaches 0.928 and both shapes reach 0.940 at $\tau{=}0.5$. On Llama-3.1-8B, saturating clipping also improves AVG by $+0.060$ over hard clipping (single run). The oracle also bounds what repair can achieve: extended runs plateau at $+0.02$ AVG over the base model, far below \ourbest, and on top of \ourbest{} the method is safe but not additive. Privileged self-distillation, even repaired, is a targeted correction tool in this regime, not a substitute for direct supervision \cite{hubotter2026sdpo,kaur2026rethinking}.

\begin{figure}[htbp]
\centering
\includegraphics[width=\columnwidth]{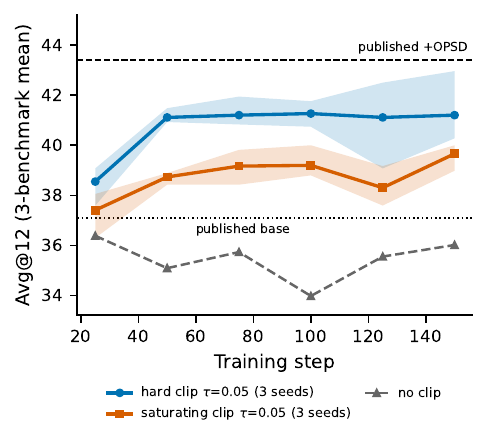}
\caption{Replication of OPSD on the original benchmarks (Qwen3-1.7B, authors' code and data). Bands show three seeds; reference lines mark the published base and best results.}
\label{fig:replcurves}
\end{figure}

\begin{figure*}[t]
\centering
\includegraphics[width=\textwidth]{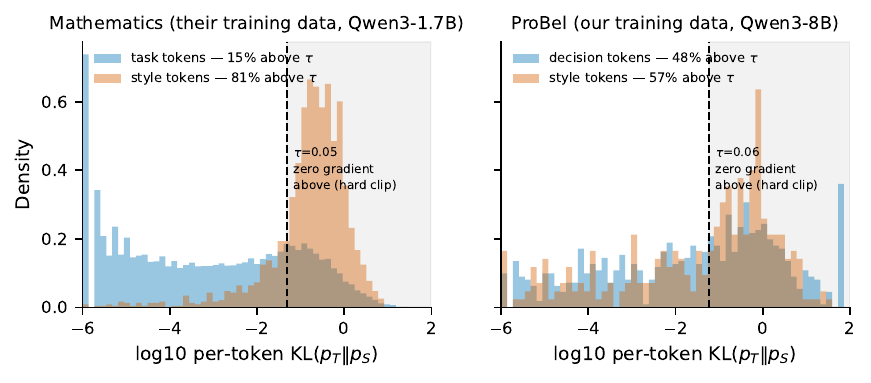}
\caption{Per-token forward KL between the privileged teacher and the student at initialization, on each domain's OPSD training data (mathematics: Qwen3-1.7B; \dsname{}: Qwen3-8B). The shaded region above the published $\tau$ receives no gradient under the hard clip. The above-$\tau$ tail is mainly stylistic on mathematics but consists of decision tokens on \dsname{}.}
\label{fig:kldomains}
\end{figure*}

\paragraph{Replication on the original benchmarks.} We replicate the published setup with the authors' code and data, Qwen3-1.7B, and evaluation protocol (Avg@12 on AIME24, AIME25, and HMMT25; best checkpoint within 100 steps). The recipe improves our base model from 34.2 to 41.9, matching the published gain (37.1 to 43.4); absolute scores are uniformly lower, including zero-shot, so we compare changes relative to our baseline. Removing the clip reproduces the published degradation (36.4 at best), while saturating clipping reaches 39.8, below hard clipping (Figure~\ref{fig:replcurves}). The gap is stable across three seeds (41.5$\pm$0.6 vs.\ 39.6$\pm$0.6), and lowering $\tau$ does not close it. The divergence profiles explain the contrast (Figure~\ref{fig:kldomains}): on mathematics, the above-$\tau$ tail is dominated by stylistic tokens (81\% exceed $\tau$), so restoring their gradients hurts; on \dsname{}, it is dominated by decision tokens (48\% above $\tau$, mean divergence $68\tau$), so hard clipping removes task supervision. Thus, the published clip is appropriate when the tail is mostly stylistic, whereas a gradient-preserving clip is needed when it carries the task signal.

\section{Significance Tests}
\label{app:significance}

Since all systems are evaluated on identical test items, paired significance testing is possible wherever prediction files exist. Table~\ref{tab:significance} reports exact (binomial) McNemar tests for central binary comparisons; ``best prompted'' is Llama-3.1-8B on Arabic and Qwen2.5-7B zero-shot on English, and the OPSD rows use Qwen3-8B with the saturating clip and the \ourbest{} initialization with the hard clip. Every fine-tuning gain is significant at $p{<}0.001$, and the open-versus-GPT-5 gap at $p{<}0.05$ on Arabic and $p{<}0.001$ on English. The final row quantifies the OPSD freeze more sharply than any aggregate metric: after 150 steps of training at the published settings on top of \ourbest, exactly 6 predictions improved and 6 degraded out of 1{,}326.

\begin{table}[htbp]
\centering
\footnotesize
\setlength{\tabcolsep}{1.6pt}
\scalebox{0.96}{
\begin{tabular}{@{}lrrl@{}}
\toprule
Comparison (binary task) & $b$ & $c$ & $p$ \\
\midrule
Llama-3.1-8B vs.\ GPT-5 CoT (AR)          & 251 & 197 & 0.012 \\
Qwen2.5-7B vs.\ GPT-5 CoT (EN)            & 746 & 417 & $<$0.001 \\
\ourbest{} vs.\ best prompted (AR)        & 242 & 116 & $<$0.001 \\
\ourbest{} vs.\ best prompted (EN)        & 396 & 225 & $<$0.001 \\
\ourbest{} vs.\ zero-shot (AR)            & 579 & 179 & $<$0.001 \\
ICL retrieved vs.\ random, $k{=}3$ (AR)   & 298 & 151 & $<$0.001 \\
Extended vs.\ original EN training (EN)   & 368 & 250 & $<$0.001 \\
OPSD saturating vs.\ zero-shot (EN)       & 311 & 180 & $<$0.001 \\
OPSD hard $\tau{=}.06$ vs.\ \ourbest{} (AR) & 6 & 6 & 1.0 \\
\bottomrule
\end{tabular}
}
\caption{Exact McNemar tests on paired binary predictions ($b$/$c$: items only the first/second system classifies correctly). The last row shows the OPSD freeze: only $6{+}6$ of 1{,}326 predictions change, so the trained model is statistically indistinguishable from its initialization.}
\label{tab:significance}
\end{table}

\section{Data Release}
\label{apndix:release}
Our proposed dataset will be released under the CC BY-NC-SA 4.0 (Creative Commons Attribution-NonCommercial-ShareAlike 4.0 International) License: \url{https://creativecommons.org/licenses/by-nc-sa/4.0/}.

% \section{Example Appendix}
% \label{sec:appendix}

% This is an appendix.

% \appendix
% \input{sections/appendix}

% \appendix
% \input{sections/appendix_v2}

\end{document}